\documentclass[9pt]{IEEEtran}
\IEEEoverridecommandlockouts
\usepackage{cite}
\usepackage{amsmath,amssymb,amsfonts}
\usepackage{algorithmic}
\usepackage{graphicx}

\usepackage{textcomp}
\usepackage{multirow}
\usepackage{framed}
\usepackage[normalem]{ulem}
\useunder{\uline}{\ul}{}
\usepackage[colorlinks, linkcolor=blue]{hyperref}
\usepackage{wrapfig}
\usepackage[numbers,sort&compress]{natbib}
\usepackage[table,xcdraw]{xcolor}
\def\BibTeX{{\rm B\kern-.05em{\sc i\kern-.025em b}\kern-.08em
T\kern-.1667em\lower.7ex\hbox{E}\kern-.125emX}}
\begin{document}
%chatgpt eda离我们有多远？
\title{ChipGPT: How far are we from natural language hardware design}

\author{
    \IEEEauthorblockN{Kaiyan Chang\IEEEauthorrefmark{1,}\IEEEauthorrefmark{2}, Ying Wang\IEEEauthorrefmark{1,}\IEEEauthorrefmark{4}, Haimeng Ren\IEEEauthorrefmark{3}, Mengdi Wang\IEEEauthorrefmark{1,}\IEEEauthorrefmark{2}, Shengwen Liang\IEEEauthorrefmark{1}, Yinhe Han\IEEEauthorrefmark{1}, Huawei Li\IEEEauthorrefmark{1}, Xiaowei Li\IEEEauthorrefmark{1}}\\
    \IEEEauthorblockA{ State Key Lab of Processors, Institute of Computing Technology, Chinese Academy of Sciences, Beijing, China\IEEEauthorrefmark{1}}
    \IEEEauthorblockA{ \\
     University of Chinese Academy of Sciences\IEEEauthorrefmark{2}\\
     School of Information Science and Technology, ShanghaiTech University, Shanghai, China\IEEEauthorrefmark{3}\\
     Corresponding Author\IEEEauthorrefmark{4}\\
   changkaiyan@live.com, wangying2009@ict.ac.cn, rhm141246718@gmail.com, \{ wangmengdi17s, liangshengwen, yinhes, lihuawei, lxw\}@ict.ac.cn}
    % Kaiyan Chang\\
    % School of Huaibei, Huaibeizhuang University, Huaibei, China

}

% \author{\IEEEauthorblockN{Yintao He\IEEEauthorrefmark{1,}\IEEEauthorrefmark{2}, Ying Wang\IEEEauthorrefmark{1,}\IEEEauthorrefmark{2}, Cheng Liu\IEEEauthorrefmark{1,}\IEEEauthorrefmark{2}, Huawei Li\IEEEauthorrefmark{1,}\IEEEauthorrefmark{2,}\IEEEauthorrefmark{3}, Xiaowei Li\IEEEauthorrefmark{1,}\IEEEauthorrefmark{2}}
%     \IEEEauthorblockA{\textit{SKLCA, Institute of Computing Technology, Chinese Academy of Sciences, Beijing, China\IEEEauthorrefmark{1}}\\
%         \textit{University of Chinese Academy of Sciences, Beijing, China\IEEEauthorrefmark{2}, Peng Cheng Laboratory, Shenzhen, China\IEEEauthorrefmark{3}}}
%     \{heyintao19z, wangying2009, liucheng, lihuawei, lxw\}@ict.ac.cn\IEEEauthorrefmark{1}}

\maketitle

\begin{abstract}
    As large language models (LLMs) like ChatGPT exhibited unprecedented machine intelligence, it also shows great performance in assisting hardware engineers to realize higher-efficiency logic design via natural language interaction. To estimate the potential of the hardware design process assisted by LLMs, this work attempts to demonstrate an automated design environment that explores LLMs to generate hardware logic designs from natural language specifications. To realize a more accessible and efficient chip development flow, we present a scalable four-stage zero-code logic design framework based on LLMs without retraining or finetuning. At first, the demo, ChipGPT, begins by generating prompts for the LLM, which then produces initial Verilog programs. Second, an output manager corrects and optimizes these programs before collecting them into the final design space. Eventually, ChipGPT will search through this space to select the optimal design under the target metrics.
    The evaluation sheds some light on whether LLMs can generate correct and complete hardware logic designs described by natural language for some specifications. It is shown that ChipGPT improves programmability, and controllability, and shows broader design optimization space compared to prior work and native LLMs alone.

\end{abstract}


\section{Introduction}

While logic design plays an essential role in agile frontend chip design,
existing manual methods require considerable effort and call for more agile
development flow. In response, trends point toward higher-level programming
interfaces like Scala and C to enhance design accessibility and efficiency. For
example, Chisel\cite{chisel} and Xilinx HLS\cite{hls} enable specification in
common languages. However, it is believed that the ultimate goal is probably
natural language logic design, where designers articulate requirements in
simple sentences. This could revolutionize chip design by maximizing creativity
and complexity at scale. Though still emerging, higher-level programming
interfaces represent progress toward this vision.
%scene

\begin{figure}[htbp]
    \centering
    \includegraphics[width=\columnwidth]{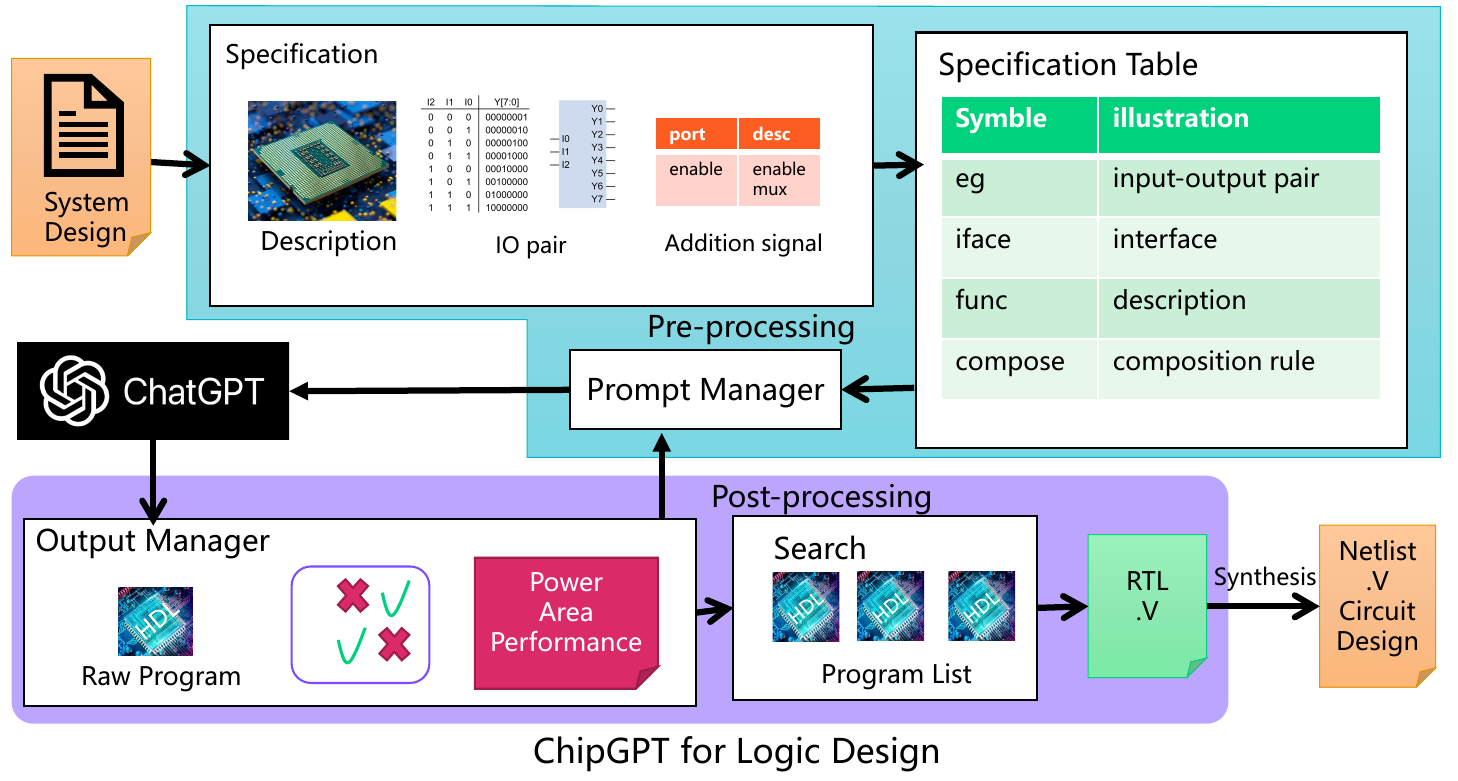}
    \caption{Integration of our ChipGPT framework to chip design flow}
    \label{fig:designflow}
\end{figure}
\begin{table*}[htbp]
    \label{tab:designmethodcomparasion}
    \caption{Agile Chip Design Methods Comparasion}
    \resizebox{2.05\columnwidth}{!}{
        \begin{tabular}{lcccccccccc}
            \rowcolor[HTML]{00BAC8}
                                                                               & ChipGPT                                                          & HLS-CPP\cite{hls}                                                     & Chisel\cite{chisel}                                             & SuSy \cite{susy}                                            & Spatial\cite{spatial} & ScaleHLS\cite{scalehls}                                                                       & TVM-VTA\cite{tvm}                                                 & CIRCT\cite{circt}                                               & XLS\cite{xls}                                                   & BoSy\cite{bosy}                                                 \\
            \rowcolor[HTML]{CBE7EB}
            \begin{tabular}[c]{@{}l@{}}SD-Algorithm \\ Simulation\end{tabular} & \multicolumn{10}{c}{\cellcolor[HTML]{CBE7EB}Python/C++}                                                                                                                                                                                                                                                                                                                                                                                                                                                                                                                                                                                                                    \\
            \rowcolor[HTML]{E7F3F5}
            \begin{tabular}[c]{@{}l@{}}SD-System   \\ Design\end{tabular}      & \begin{tabular}[c]{@{}c@{}}Decomposed  \\  Document\end{tabular} & Document                                                              & Document                                                        & Document                                                    & Document              & \begin{tabular}[c]{@{}c@{}}Document/DL  \\  Model\end{tabular}                                & DL   Model                                                        & Document                                                        & Document                                                        & Document                                                        \\
            \rowcolor[HTML]{CBE7EB}
            \begin{tabular}[c]{@{}l@{}}SD–Specialize \\ Field\end{tabular}     & \begin{tabular}[c]{@{}c@{}}General   \\ Digital IC\end{tabular}  & \begin{tabular}[c]{@{}c@{}}Algorithm   for \\ Digital IC\end{tabular} & \begin{tabular}[c]{@{}c@{}}General  \\  Digital IC\end{tabular} & \begin{tabular}[c]{@{}c@{}}Systolic  \\  Array\end{tabular} & CGRA/FPGA             & \begin{tabular}[c]{@{}c@{}}Algorithm   for \\ Digital IC/\\ DL Model\end{tabular} & \begin{tabular}[c]{@{}c@{}}DL   Model \\ Accelerator\end{tabular} & \begin{tabular}[c]{@{}c@{}}General  \\  Digital IC\end{tabular} & \begin{tabular}[c]{@{}c@{}}General  \\  Digital IC\end{tabular} & \begin{tabular}[c]{@{}c@{}}General \\   Digital IC\end{tabular} \\
            \rowcolor[HTML]{E7F3F5}
            Logic-Module Input                                                 & Prompt                                                           & C++/LLVM                                                              & Scala/FIRRTL                                                    & URE/Halide   IR                                             & Spatial-Scala         & ONNX                                                                                          & ONNX                                                              & MLIR-Affine                                                     & Rust   like                                                     & CTL   Logic                                                     \\
            \rowcolor[HTML]{CBE7EB}
            Logic-Module Output                                                & Verilog                                                          & Verilog                                                               & Verilog                                                         & HLS-C++                                                     & Verilog               & Verilog                                                                                       & Pynq                                                              & Verilog                                                         & Verilog                                                         & Verilog                                                         \\
            \rowcolor[HTML]{E7F3F5}
            Logic verification                                                 & \multicolumn{10}{c}{\cellcolor[HTML]{E7F3F5}Verilator, ModelSIM}
        \end{tabular}
    }
\end{table*}%表格中的文本没有对应的引用数据

To equip logic design with a higher programming interface, researchers in the
program synthesis community provide methods that generate Verilog from formal
representation\cite{taedp,formalmother} and input-output pairs\cite{syntax},
such as Bosy\cite{bosy}. These intuitive approaches can release programmers
from writing sophisticated programs. However, it still suffers from issues of
correctness and completeness, to which the recent deep learning based code
generation model reveals a proper solution. The input to these fundamental
generative models is natural language, while the output is the target program.
For example, one of SOTA models for this case is ChatGPT, which receives
natural language input and potentially facilitates natural language
programming. %related work, advantages 

While large language models (LLMs) show promise as universal generators, they
face limitations in adapting to chip design. Their ambiguous input and output
prevent seamless integration into electronic design automation (EDA) workflows.
From an input perspective, prompt engineering for hardware description language
(HDL) generation remains unrefined. First, LLMs lack prompt templates producing
high-quality HDL code. Second, they cannot generate customized hardware modules
or apply top-down design principles. From the output perspective, LLMs only
generate raw programs which neither guarantee hardware-level correctness nor
enable PPA exploration in the potential design space.  %related works limitations, challenge, observation

From the toolset perspective, large foundational models often run on cloud
servers, preventing fine-tuning for logic design\cite{ict,lora}. To overcome
the challenges and limitations above, we propose a natural language chip logic
design method based on in-context learning, which does not modify the large
model itself. As shown in Fig. \ref{fig:designflow}, the core of our method is
to put a prompt manager in front of the GPT model, which helps designers to
improve the program quality by generating high-quality prompts. Secondly, to
enable the PPA optimization process of design generation, we set an output
manager behind GPT. The output manager corrects the initial programs through
machine feedback and human feedback and then collects these raw programs into a
potential program list. Afterward, an enumerative search stage is invoked to
select the best design(\emph{i.e.} with target PPA) from the generated program
list. The evaluation results show that our framework can improve the average
effect when dealing with exemplary design cases. %method
% The detailed experimental results are in \href{https://anonymous.4open.science/r/chipgptresult-C12E}{https://anonymous.4open.science/r/chipgptresult-C12E}. %result(need to update)
% (deleted) in our evaluation section 

%contribution
ChipGPT is an attempt that explores and estimates the feasibility of
automatically generating logic design using natural language chip
specification, and it makes use of current LLMs to ease the cost of hardware
frontend design, which traditionally requires a high degree of expertise and
manual labor. The contributions are listed below:
\begin{itemize}

    \item We are the first to evaluate the ability of large language models(\emph{i.e.}
          ChatGPT) to generate hardware logic design from natural language specification.
          To interface the LLM to hardware designers, we also propose a scalable
          four-stage zero-code logic design framework based on GPT. This framework
          facilitates natural language-based chip design starting from the chip
          specification, thereby reducing manual design effort.
    \item To improve the quality of logic design, we are the first to propose a portable
          framework to generate Verilog program without retraining or modifying any
          weights in LLM(\emph{e.g.} fine-tuning, adapter layer), which can be seamlessly
          integrated into the latest LLM APIs.
    \item We overcome the challenges brought by the power/area/performance agnostic LLMs
          on the chip design process through inserting a post-LLM search method with an
          efficient LLM output manager.
    \item Compared with previous agile chip logic design methods and the native ChatGPT,
          ChipGPT shows improvement in the programmability and scalability aspect, which
          shows potential to be extended to larger-scale chip design.
          % \item ChipGPT can be practical in use as a hardware development tool, which we implement as a javascript plugin in html for logic design.

\end{itemize}
\section{Background and Motivation}
\subsection{Agile Hardware Design Workflow}

Agile chip design approaches fall into two categories: programming
language-based and program synthesis-based, as shown in Table
\ref{tab:agiledesignworkflow}. Programming language-based methods aim to
increase productivity by using higher-level languages\cite{susy}. For example,
high-level synthesis (HLS) is the process of automatically generating a
register-transfer level (RTL) hardware description from a high-level
programming language(\emph{e.g.} C/C++). Chisel is a Scala-embedded
domain-specific language helping designers efficiently produce RTL code.
However, these programming language-based agile chip design flows do not accept
natural expression, which still needs hand-crafted design programming.% HLS, Chisel traditional agile design workflow

Program synthesis automatically generates computer programs from high-level
descriptions like specifications, examples or input-output pairs. This enables
faster, less error-prone program development than manual methods. Two
approaches are inductive and deductive synthesis. Inductive synthesis uses
input-output examples, iteratively refining a program to satisfy
them\cite{iosynthesis}, which is not always completely correct and rarely used
in accurate hardware design. Another synthesis method is called deductive
synthesis, also known as formal methods, and it is a process of synthesizing
correct-by-construction hardware designs using mathematical models and formal
verification techniques\cite{bosy,boundsynthesis}. This method has been used in
Verilog generating(\emph{e.g.} Bosy) in prior works. However, this process
entails proving the correctness (soundness) of the design before
implementation, thereby raising design difficulty. In general, these program
synthesis-based agile workflows rely on the quality of complete examples
provided or a formal specification that is hard to understand and learn.% program synthesis agile design workflow

To compare previous agile design methods with our trial of LLM-based natural
language hardware design flow, ChipGPT, Tab. \ref{tab:designmethodcomparasion}
lists several common agile chip design methods. Columns represent the EDA
workflow from system-level to logic design. First, algorithm simulation in
Python or C++ checks design correctness. In the system design stage(\emph{i.e.}
2th row), all methods follow design documents except TVM-VTA which uses a
specialized deep learning model description format. The "specialized
field"(\emph{i.e.} 3th row) denotes whether a method is general purpose or
domain-specific. In logic design, different input representations indicate
unique productivities. Since the input of ChipGPT is natural language, ChipGPT
is the most efficient representation compared with other methods.
\begin{figure*}[] %通栏
    \begin{minipage}[t]{0.33\linewidth} %调节两个子图左右间距
        \centering
        \includegraphics[width=1.0\linewidth]{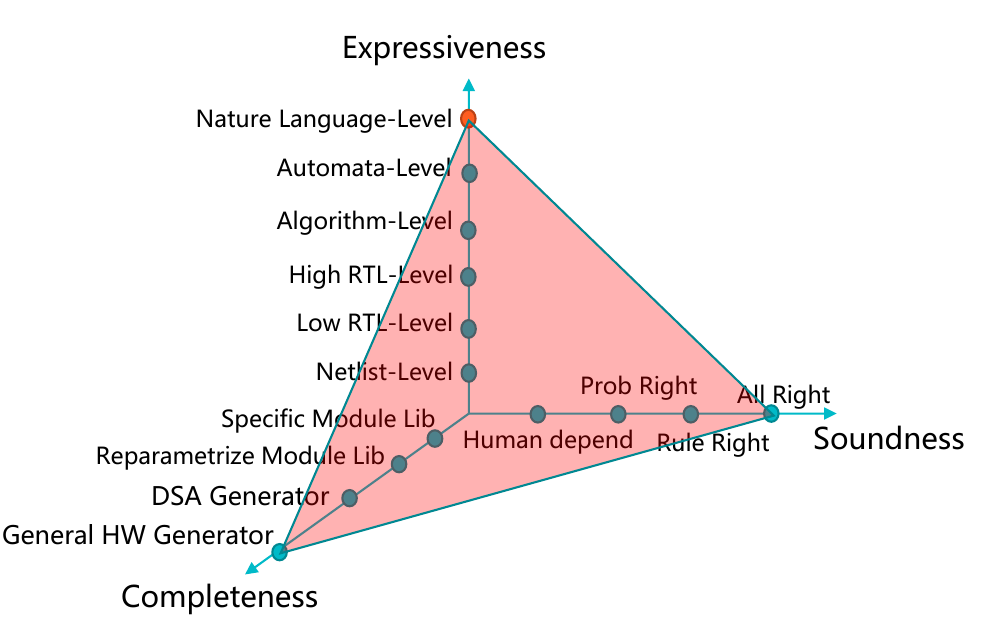} %调节单个子图大小
        \caption{Ideal Agile method productivity} %子图下标题
        \label{fig:idealagile} %引用标签
    \end{minipage}%
    \begin{minipage}[t]{0.33\linewidth}
        \centering
        \includegraphics[width=1.05\linewidth]{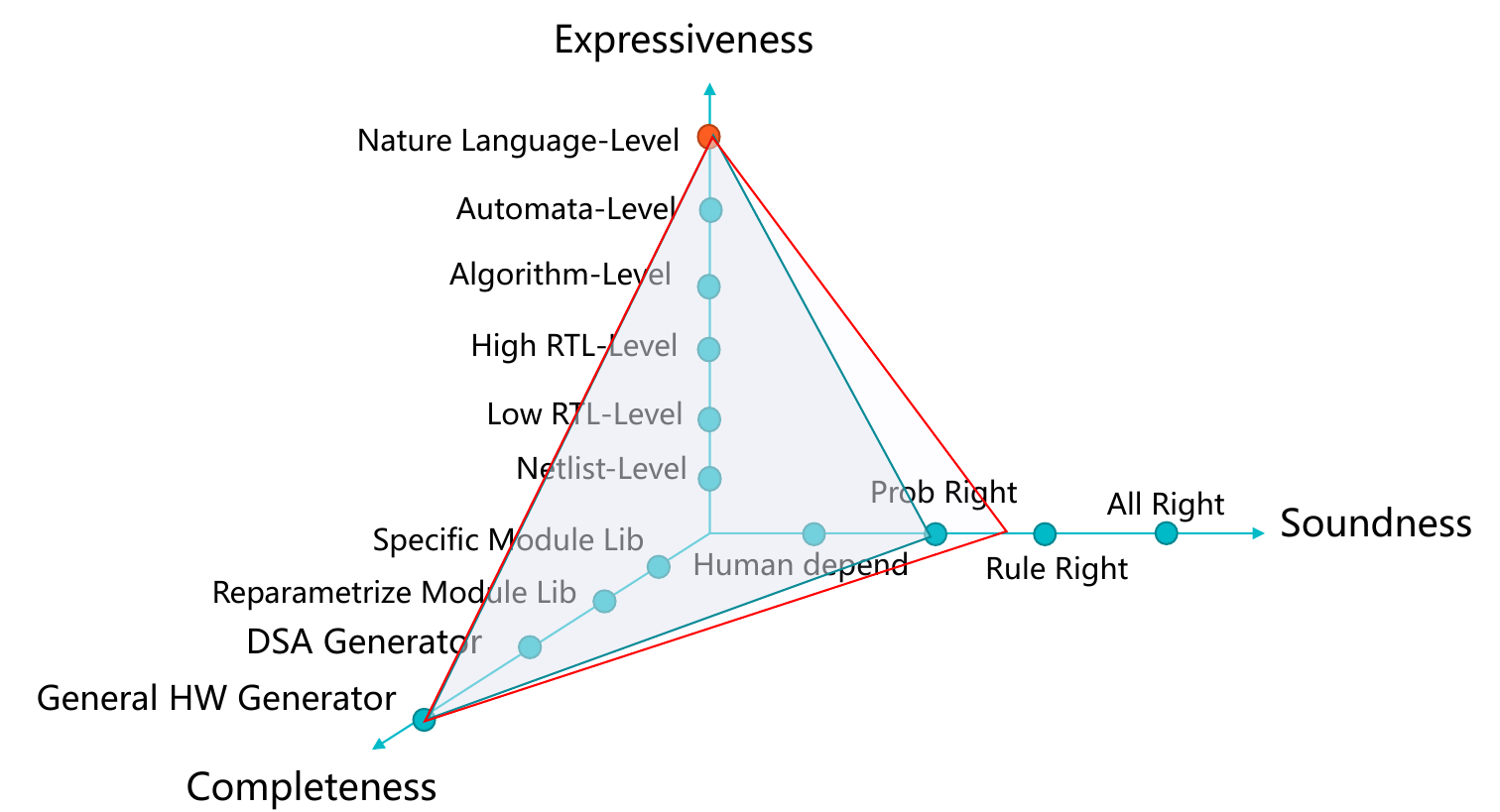}
        \caption{ChipGPT productivity improvement}
        \label{fig:chipprodimprv}
    \end{minipage}%
    \begin{minipage}[t]{0.33\linewidth}
        \centering
        \includegraphics[width=0.93\linewidth]{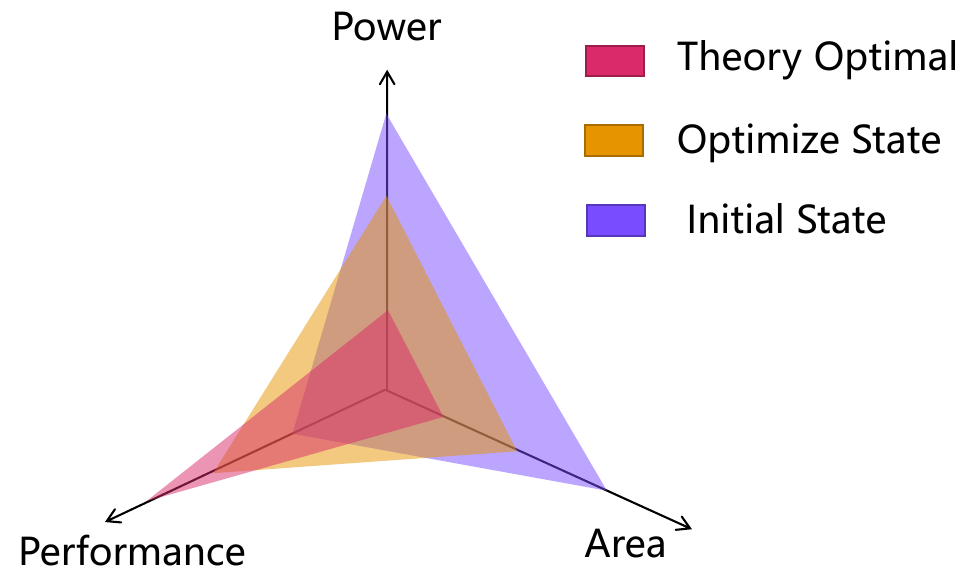}
        \caption{ChipGPT design space exploration}
        \label{fig:chipdse}
    \end{minipage}
\end{figure*}

\begin{figure}
    \centering
    \includegraphics[width=\columnwidth]{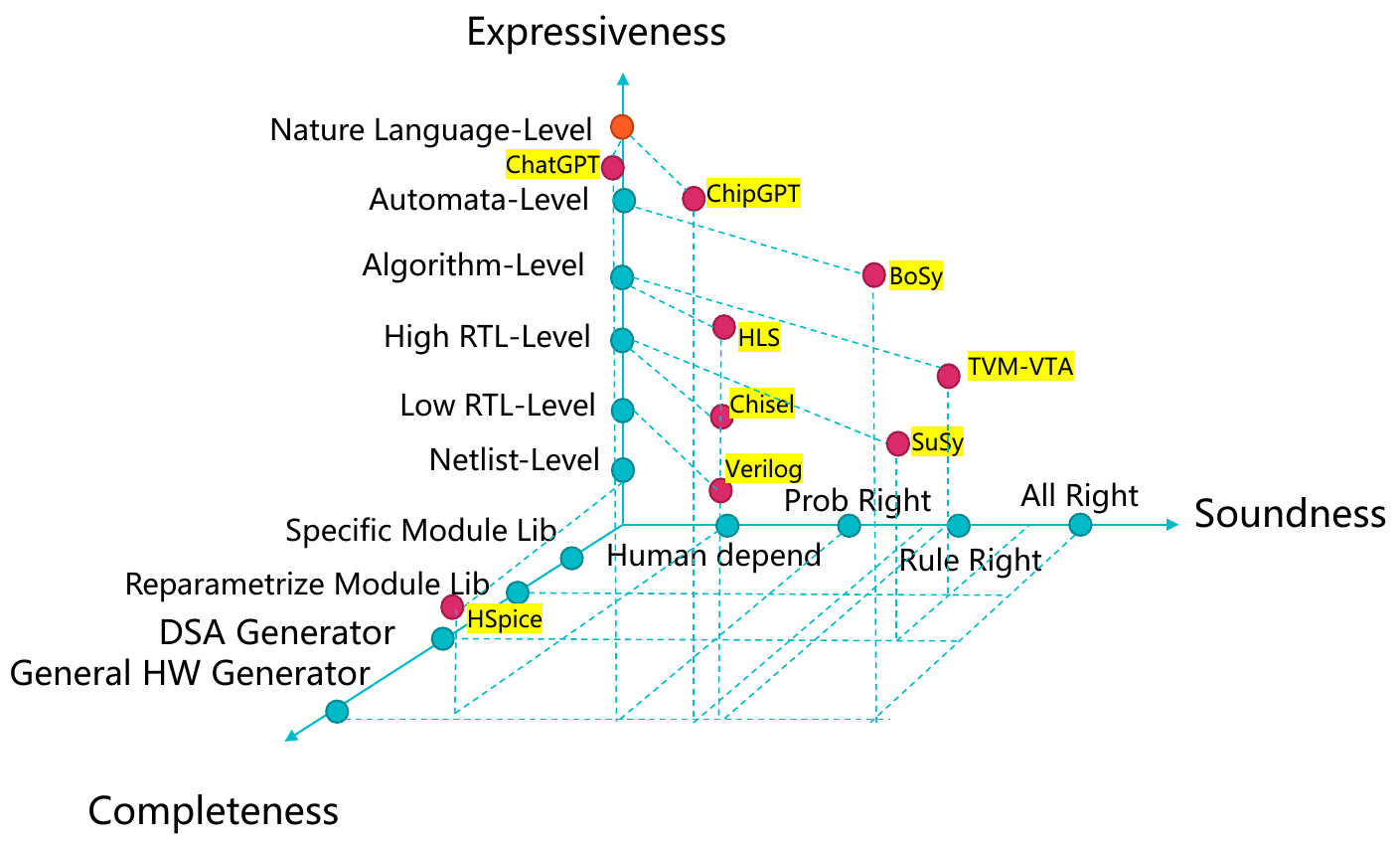}
    \caption{Productivity Measurement of Agile Chip Design Method}
    \label{fig:agilemeasure}
\end{figure}

\begin{table}[htbp]
    \label{tab:agiledesignworkflow}
    \centering
    \caption{The development of Agile hardware design}
    \begin{tabular}{ll}
        \rowcolor[HTML]{00BAC8}
        \textcolor{white}{Agile Logic Design Method     }                                        & \textcolor{white}{Representative Works }                                                                                          \\
        \rowcolor[HTML]{CBE7EB}
        \begin{tabular}[c]{@{}l@{}}Programming   language based\\   Hardware design\end{tabular} & \begin{tabular}[c]{@{}l@{}}Chisel\cite{chisel},   Spatial\cite{spatial}, \\ SuSy\cite{susy}, ScaleHLS\cite{scalehls}\end{tabular} \\
        \rowcolor[HTML]{E7F3F5}
        \begin{tabular}[c]{@{}l@{}}Program   Synthesis based\\  Hardware design\end{tabular}     & BoSy\cite{bosy},   Bounded Synthesis\cite{boundsynthesis}
    \end{tabular}
\end{table}
\subsection{Agile Design Flow Comparison}\label{sec:agilemeas}
To compare agile chip design productivity, we propose a 3D measurement space
(Fig. \ref{fig:agilemeasure}). The first axis is soundness, indicating
generated hardware description correctness. For example, BoSy uses formal
methods, producing fully correct programs. ChatGPT generates correct programs
probabilistically (prob-right). Methods like HLS and Chisel produce correct
programs if rules are followed (rule-right). The second axis is completeness,
denoting how much of the design space a method covers. For example, TVM-VTA
generates domain-specific architectures (DSA generator), covering part of the
space. General methods like HLS and Chisel can describe most logic structures
(General HW Generators). The third axis is expressiveness, indicating input
language productivity. For example, Verilog has RTL-level expression (Low
RTL-level). HLS uses algorithmic languages like C (algorithm-level). ChatGPT
and ChipGPT use natural language (highest expressiveness). The inputs of
ChatGPT and ChipGPT are all-natural language descriptions, thus they are all at
the highest expressiveness level.

\subsection{Transfer learning}
%why we choose in-context learning
Large language models have become the foundation of code generation tasks, as
Generative Pre-Trained Transformer (GPT) benefits from abundant global data
(\emph{ChatGPT}). It shows advantages in generating
programs\cite{codebert,codet5,graphcodebert,unixcoder}. To apply large language
models to the chip design field, we explore several common transfer learning
methods(\emph{i.e.} fine-tuning\cite{finetuning}, adapter\cite{adapter},
lora\cite{lora}, in-context learning\cite{ict}) and choose in-context learning
as the final solution. This is because fine-tuning involves retraining
pre-trained language model on task-specific data which is unrealistic since LLM
serves in cloud and does not permit users' retraining. Adapter method and lora
require additional layers to train, which requires repeatedly querying the LLM
API, leading to inefficiency and cost. Therefore, in-context learning appears
as an emerging prompt tuning approach and is widely used in the natural
language processing community.

%What is ICL
In-context learning (ICL) elicits responses from a pre-trained LLM without
retraining, directly predicting results based on prompts in natural language
context. The key observation of ICL is that an LLM's responses depend on all
preceding prompts. It incorporates a prompt manager generating prompts that
guide the LLM's answers.\cite{humanprompt}.
%detail principle of in-context learning
The core of prompt manager is to generate prompts using templates, which follow
several general design principles (\emph{e.g.} chain-of-thought\cite{cot},
least-to-most\cite{leasttomost}). However, prompt managers differ across
fields(\emph{e.g.} Visual ChatGPT\cite{vchatgpt}) providing an opportunity for
transfer learning in chip design.

\subsection{Motivation}\label{sec:motivation}

Emerging agile design workflows focus on EDA front-ends with higher-level
representations. We analyzed current method productivity based on
expressiveness, soundness and completeness (Sec. \ref{sec:agilemeas}). The
ideal agile method is shown by the red triangle in Fig. \ref{fig:idealagile}.
Ideal productivity requires: 1) Natural language expression for high
abstraction and productivity. 2) A general hardware generator covering most of
the design space. 3) Guaranteed correctness of all generated hardware
descriptions.

Large language models (LLMs) enable natural language hardware descriptions,
covering expressiveness (blue area, Fig. \ref{fig:chipprodimprv}). However,
LLM-generated programs may lack correctness (soundness). Enhancing correctness
is key to applying LLMs for chip design, presenting three primary challenges:

\paragraph{Challenge 1} \textbf{LLM inputs are ambiguous, unable to directly integrate into chip design flows.} LLMs receive query sequences and output answer sequences. Producing HDL requires queries from specifications, which are complex with many module descriptions. Determining prompts eliciting a desired module is difficult. Therefore, we propose a specification split method in Sec. \ref{sec:specsplit} and template-based prompt manager \ref{sec:prommanager} to overcome this challenge. These improve soundness from probable to nearly rule-right (red area, Fig. \ref{fig:chipprodimprv}).  %解释为什么大模型需要formal的表示方法

\paragraph{Challenge 2} \textbf{LLMs are unaware of power, area and performance (PPA), unable to generate ideal programs (red, Fig. \ref{fig:chipdse}). }  As shown in Fig. \ref{fig:chipdse}, the ideal program PPA results are in the red area, but LLM-generated programs remain in the initial state (PPA-agnostic) due to their training. LLMs like InstructGPT and ChatGPT are trained using general reinforcement learning with a generic reward function to improve performance. This results in programs that are intuitively good but not optimized for PPA. Therefore, we propose an output manager in Sec. \ref{sec:outputmanager} to tackle this challenge and promote the PPA to the yellow area in Fig. \ref{fig:chipdse}.% why the challenge exists, has what results.

\paragraph{Challenge 3} \textbf{It cannot expand to unlimitedly nested architecture.} LLM-generated programs consist of known modules, limiting scalability to generate top modules with customized submodules.  Therefore, we propose a bottom-top method to overcome this challenge in Sec. \ref{sec:comp}.
%使用principle来解释代码生成，而不是原有的
\section{Automatic Chip Generation Framework}

ChipGPT serves as an EDA frontend framework, which aims to assist humans in
compiling chip specifications to logic design. It takes chip specification as
input and generates target hardware module description (\emph{i.e.} Verilog
program). %provide the research goal of the framework

$$HDL=ChipGPT(specification)$$

\begin{figure}
    \centering
    \includegraphics[width=\linewidth]{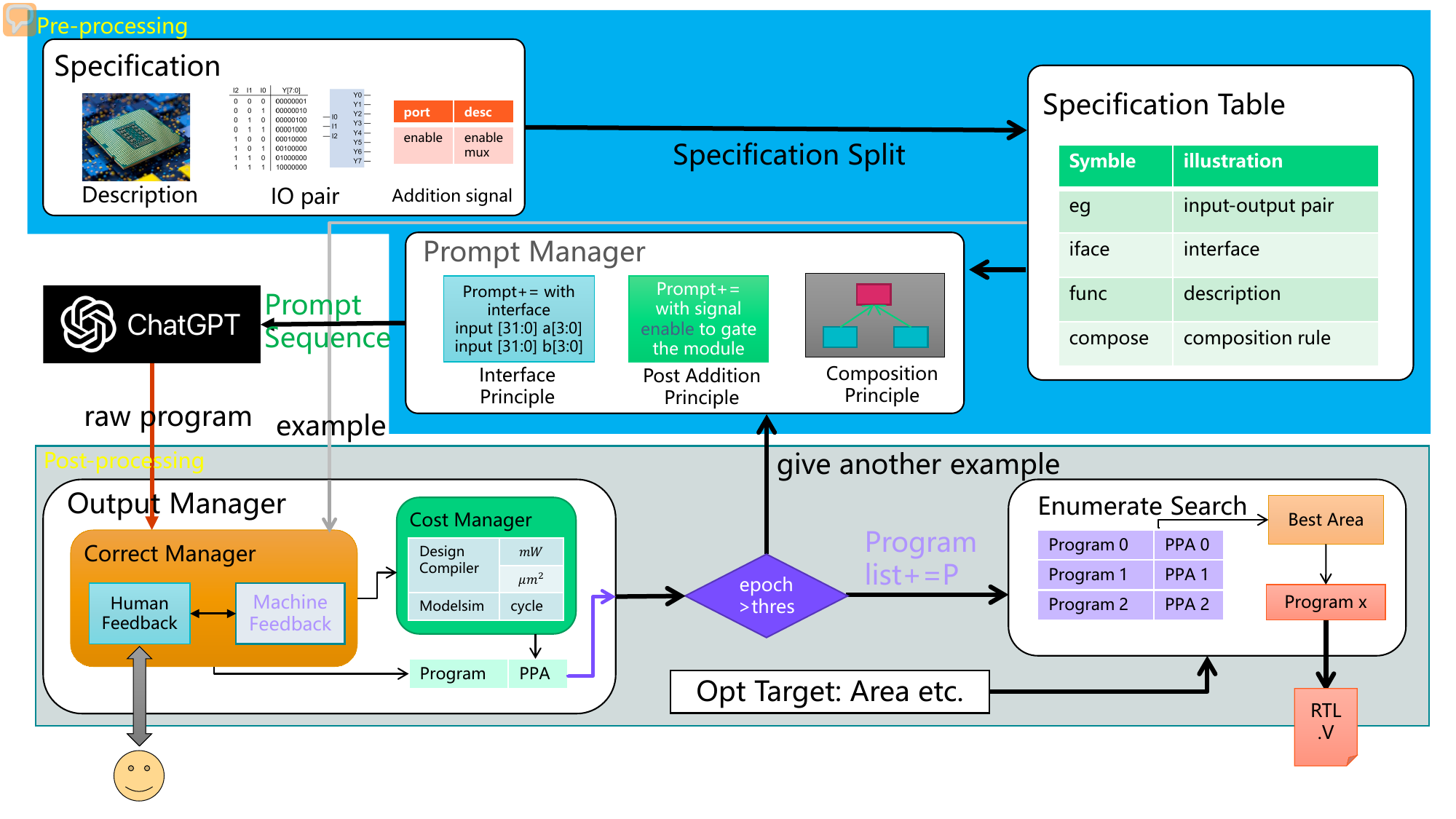}
    \caption{The ChipGPT framework overview. Specification Split and Prompt Manager implicitly control the output (Pre-processing). Output Manager and Enumerative Search explicitly control the output (Post-processing).}
    \label{fig:overview}
\end{figure}
%illustrate the variables above

The chip specification defines the target of HDL implementation. Thus HDL
should be controlled by chip specification. To control the generated HDL, there
are three possible solutions in the GPT-based code generation workflow. 1)
Control the context by customizing configuration prompts(\emph{i.e.} context).
2) Control the code query prompt by modifying the generated prompt. 3) Directly
modify the generated program. Among them, 1) and 2) implicitly control output
by changing the queries in the dialogue sequence $S$. 3) is an explicit control
method via program analysis. Due to the GPT-produced program not always being
accurate as well as challenging to analyze and modify, ChipGPT does not attempt
to amend the generated raw program directly. Rather, it gives GPT several times
of trials to generate different code versions and refine them to obtain the
final output. The workflow under this design consideration is shown in Fig.
\ref{fig:overview}. Specification Split and Prompt Manager implicitly control
the output. Output Manager and Enumerate Search explicitly control the output.%overview description and the design considerations

At each iteration, ChipGPT follows these steps. At first, pre-processing occurs
before the GPT model is invoked to handle descriptive and irregular information
(\emph{e.g.} the function of a module and interface descriptions).
Post-processing then takes place after the GPT model. The output manager and
search modules are used in post-processing to address design constraints
(\emph{e.g.} regardless of whether the raw program is correct, ChipGPT will
select a version with optimized power, performance, and area). $$split(spec) =
    \{eg,iface, func, compose\}$$ $$rawcode=GPT(PM(iface,func,compose,fd))$$
$$code=OM(rawcode,EM)$$ $$codelist=codelist \cup code$$
$$optcode=Search(codelist,requirements)$$ %formalize description

First, to translate the specifications into a formal and unambiguous form, we
propose a specification split method. %why take spefication as input

Second, to make these irregular split partitions become an available GPT input
sequence, we propose a specification serialization method in $PM$. % why should have prompt manager

Third, to improve the quality of the $code$(raw program) generated by the GPT
model, we propose an output manager $OM$.% why should have output manager

Finally, the code generated in the third stage typically does not meet
performance, power and area (PPA) requirements. To obtain an optimized final
output program for PPA, we propose a search method compatible with the previous
stages that ranks GPT-generated programs and identifies the version according
to the $requirements$. % why should have search stage

\subsection{Specification Split}\label{sec:specsplit}

A comprehensive chip logic design specification typically comprises module,
test and interface descriptions that provide the necessary information to
translate into functionally verified hardware implementations. To enable
hardware developers to precisely understand module functionality,
specifications often provide input-output examples for the interfaces. %sepcification definition

Based on the above specification definitions, we manually divide the
specification into four parts (\emph{i.e.} module descriptions, test
descriptions, interface descriptions, input-output examples ). We discard any
other extraneous parts of the specifications.

\begin{equation}\label{rqu:split}
    split(spec) = \{eg,iface, func, compose\}
\end{equation}

$eg$ represents examples within the module specifications. This information is often provided in tables where each row lists inputs and outputs within a clock cycle.  ChipGPT does not take $eg$ as the prompt input, for the reason that this information is too complicated to fit the transformer model. Rather, we handle it in the post-processing step. The Output manager leverages this information to verify module correctness, reducing the need for manual human correction.

$iface$ represents the input and output ports definition.

$func$ represents the function of the module, which is the most crucial part of the ChipGPT workflow. The module function defines its core purpose and behavior.

$compose$ delineates how to compose the module, as designated when the module comprises the top level of the design.

\subsection{Prompt Manager $PM$}\label{sec:prommanager}
Due to the context-aware feature of GPT, the prompt manager directly determines
the quality of the generated code.

The goal of the prompt manager is to serialize natural language software
specifications into a structured prompt format. By formatting the
specifications this way, the prompts can serve as input for GPT. With
well-designed prompts, GPT can produce code that successfully captures the
details and purpose of the original specifications.

\begin{equation}\label{equ:arch}
    arch = <setup,submodule^N, compose>
\end{equation}
\begin{equation}\label{equ:funcsplit}
    submodule=<\{func desc, iface\},addition>
\end{equation}

To improve prompt quality, we design a template-based prompt manager where
prompts are interconnected within templates. We first provide a formal
representation for the prompt manager. Let $S=\{(Q_1, A_1),(Q_2,A_2),\dots ,
    (Q_N,A_N)\}$ denote a dialogue sequence with $N$ question-answer pairs. GPT
takes queries as input and generates the final answer. We define $arch$ in Equ.
\ref{equ:arch} as a query sequence, or prompt sequence illustrated in Fig.
\ref{fig:overview}. The query sequence comprises three partitions. Elements
enclosed in $<>$ represent a sequence of prompts, while elements in $\{\}$
represent groups combined into a single prompt. The first setup prompt $<$ act
as a proficient Verilog programmer$>$ establishes an execution environment
enabling the following queries to operate in a Verilog context. Next, for the
basic modules in the specification, GPT generates $N$ modules following the
interface model principle in \ref{sec:ifm} and post-addition principle in
\ref{sec:postadd}. Finally, there is a top module following the $compose$ rule
that connects these submodules from the second step. This composition adheres
to the module composition principle in \ref{sec:comp}.  %formalize ChipGPT model and prompt principle insight

\subsubsection{Principle for Interface Model}\label{sec:ifm}
\begin{figure}
    \centering
    \includegraphics[width=\linewidth]{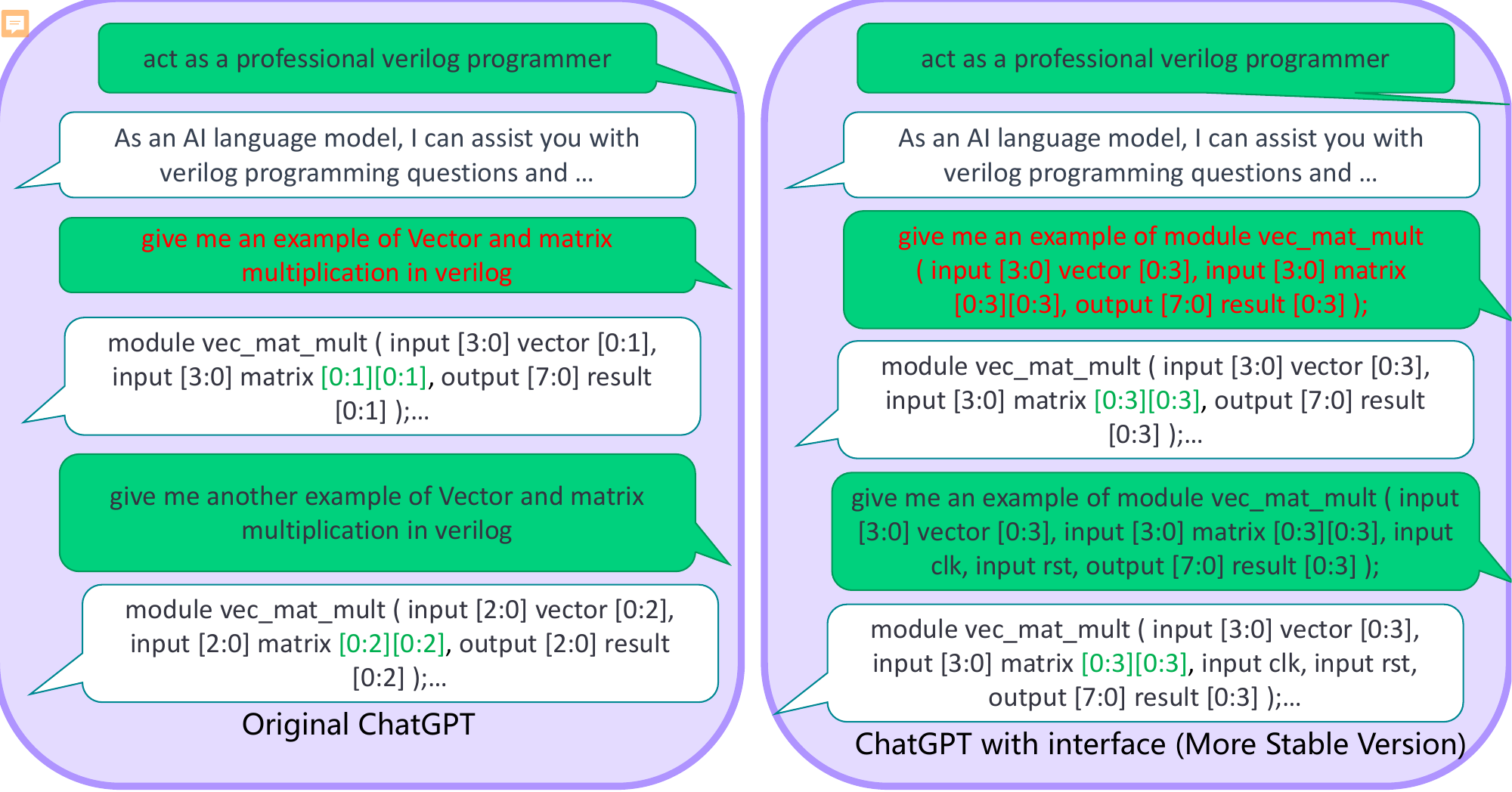}
    \caption{An observation of interface principle in prompt manager}
    \label{fig:interfaceprinciple}
\end{figure}
An intuitive approach for generating Verilog programs is to specify the module function as the prompt only. However, omitting the module interface declaration from the prompt makes the GPT ignore chip port specifications, where the generated module cannot be integrated seamlessly into the testbench environment. Moreover, including the interface declaration within the prompt defines attributes like bit widths and array sizes that are essential for generating high-quality internal module code. This enhanced context enables GPT to produce consistent versions with the same interface, as depicted in Fig. \ref{fig:interfaceprinciple}.%insight

To introduce the interface-based model principle into our template-based prompt
manager $PM$, we examine current port definitions in specifications and
hardware description languages (HDLs), such as Verilog. Port definitions
comprise five elements: 1. Direction ($D$): Whether the port is input, output,
or inout 2. Data width ($W$): The bit width of the port (e.g. 32 bits) 3.
Variable name ($VN$): The name given to the port (e.g. instruction) 4. Array
size ($S$): The number of elements in an array port 5. Port description ($PD$):
A text description of the port's purpose.

$$iface=\{ D, W, VN, S, PD\}$$

Among the port definition elements, the variable name ($VN$) is particularly
important. Because GPT is a variable name-aware model, the names given to ports
help the model infer the appropriate program structure. %action and description

\subsubsection{Post Addition Principle}\label{sec:postadd}
\begin{figure}[htbp] %通栏

    \centering
    \includegraphics[width=1.0\linewidth]{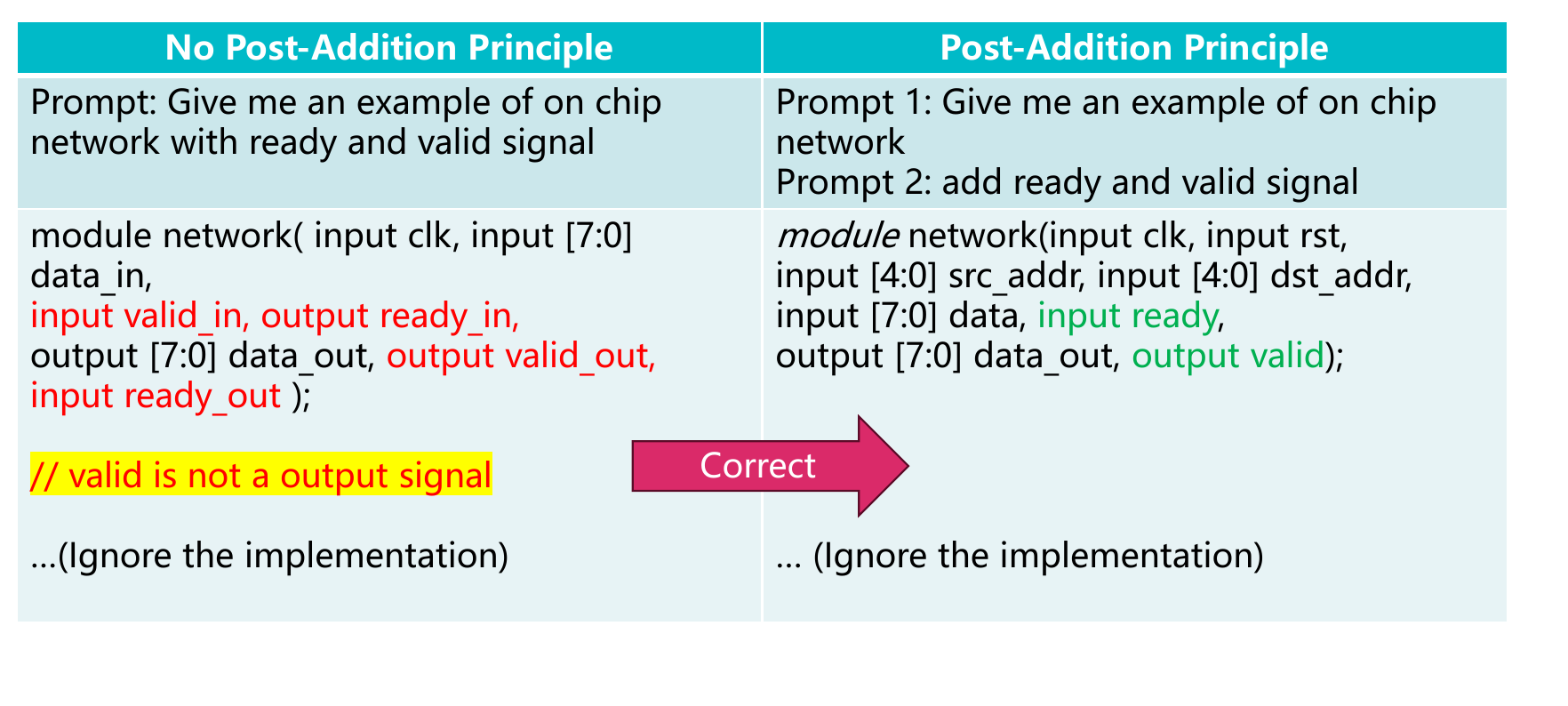} %调节单个子图大小
    \caption{An example of post-addition Principle in LLM\cite{claude}.} %子图下标题
    \label{fig:postaddition} %引用标签
\end{figure}
%需要添加例子说明observation
In a production environment, multiple cross-module handshake signals are often
added to existing modules(\emph{e.g.} ready-valid mode). However, composing
handshake signals directly into the interface prompt reduces the accuracy of
the generated raw program. GPT struggles to well understand programs created
this way. The reason is that LLM's reward model always selects the
highest-ranked raw program version based on both the handshake and function. It
considers the weights of the handshake and function together. However,
programmers prioritize function correctness.  %insight

Based on the above observations, we place additional information such as
cross-module handshake signal declarations after generating the first raw
program in Equ. \ref{equ:funcsplit}. Because this supplemental information is
added following the primary module details, we call this the "post-addition
principle". For example, in Fig. \ref{fig:postaddition}, we show the
effectiveness of the post-addition principle. In the left column, the handshake
signals are added to the same prompt. As a result, the language model generates
an incorrect implementation where valid cannot be an output port. The right
column implements the post-addition principle, where the implementation is
correct.%action and description
%post-addition部分加上图，intro部分加上图，composition部分加上图，实验部分加上CPU的spec，并且加上矩阵乘法会产生性能差异的图。
\subsubsection{Module Composition Principle}\label{sec:comp}
%需要添加例子说明observation
% \begin{figure}[]
% \centering
% \includegraphics[width=1.0\linewidth]{compose1.pdf}
% \caption{An observation of defining a complex module in one prompt.}
% \label{fig:compose1}
% \end{figure}
\begin{figure}[] %通栏

    \centering
    \includegraphics[width=1.0\linewidth]{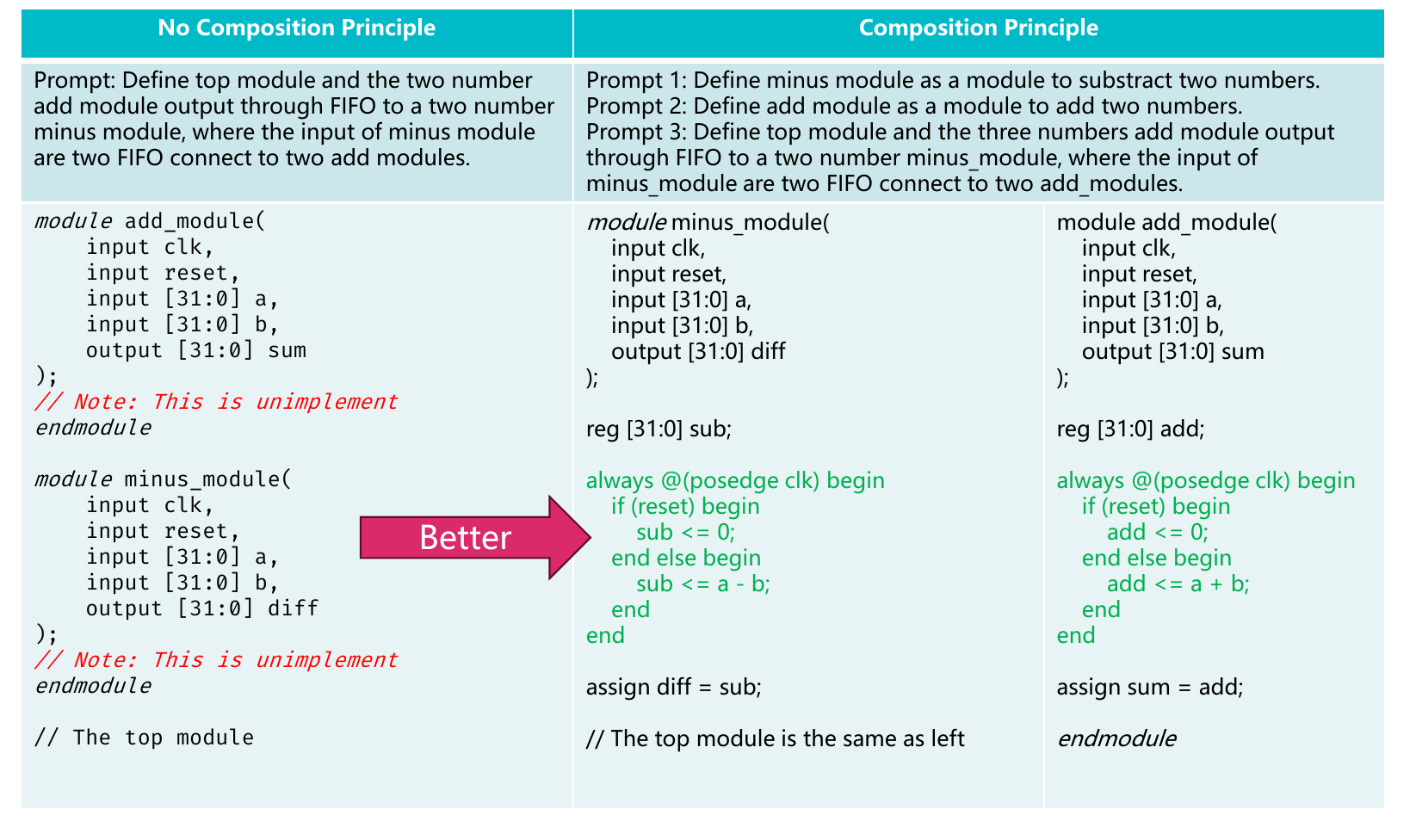} %调节单个子图大小
    \caption{An example of composition principle in LLM\cite{claude}.} %子图下标题
    \label{fig:compose} %引用标签
\end{figure}

An architecture-level chip specification includes not only component modules
but also a top module to connect them. To gain a high-level overview, chip
designers typically take a top-down design approach. However, if we prompt GPT
to generate the top-level module directly, it lacks information about the
submodules it should reference. %insight

Based on this observation, we propose the "bottom-up composition principle" to
improve the quality of the generated raw program, which is similar to least-to
most principle\cite{leasttomost}. As shown in Equation \ref{equ:compose}, this
principle specifies that \textbf{submodule interfaces must be declared before
    the top module}. For example, $<$ the submodule has interface $\dots$ $>$
followed by $<$ the top module consists of $\dots$ $>$. By following this
sequence, when GPT generates the top module, the submodules' ports are clearly
defined. As a result, GPT can connect them appropriately. Because this
principle is recursive, it can be applied to a wide range of architectural
design scenarios. For example, in Fig. \ref{fig:compose}, we want to design a
complex module. On the left, the add module and minus module are defined in the
same prompt, where the LLM cannot output the correct implementation of the
submodules. However, when applying the composition principle, the LLM can
generate correct implementations of the submodules.  %action and description

\begin{equation}\label{equ:compose}
    compose: {submodule}^N \longrightarrow topmodule
\end{equation}

\subsection{Output Manager $OM$}\label{sec:outputmanager}
The programs GPT generates (referred to as "raw" programs) may not adhere to
specifications. To address this issue, an additional correct model is needed to
improve accuracy. A cost model can then optimize the raw program for power,
performance, area (PPA) and other metrics based on human-specified targets. The
workflow incorporating these models is as follows:

\begin{equation}
    OM= Correct \hookrightarrow Cost
\end{equation}
\subsubsection{Correction Manager}
In order to correct raw program output from GPT, programmers often fix them
manually. However, we found the examples provided in the specification
facilitate machine-based feedback. Thus we propose a two-stage correction
manager. The first stage is to use a compiler and simulator to check its
correctness. If the machine feedback is correct, then the program does not need
human feedback. The second stage is to manually check the raw program. After
these steps, the raw program is corrected.
\begin{equation}
    Correct=Machine\_correct\hookrightarrow Human\_correct
\end{equation}
\paragraph{Machine Feedback} Machine feedback checks the correctness of the raw
program generated by GPT. We can use machine feedback models because
specifications provide sound examples(the examples are guaranteed to be
correct).%insight
\begin{equation}
    Machine\_correct=simulator(synthesis(raw),eg)
\end{equation}

Raw programs and the corresponding examples are fed into the simulator to test
if modules are functional right. %action
\paragraph{Human Feedback} Human feedback follows machine feedback for two reasons.
First, the machine feedback stage may become stuck at a compile error it cannot
address. Second, the examples provided by only the machine feedback stage are
not complete. Due to the above observations, programmers need to correct the
program manually. Note that this stage serves as an auxiliary for the
composition principle to reduce incorrect generation errors. However, it has
little effect within the ChipGPT automatic framework discussed further in
Section \ref{sec:eval} regarding Research Question 5. %insight

\subsubsection{Cost Manager}

\begin{figure}[] %通栏

    \centering
    \includegraphics[width=1.0\linewidth]{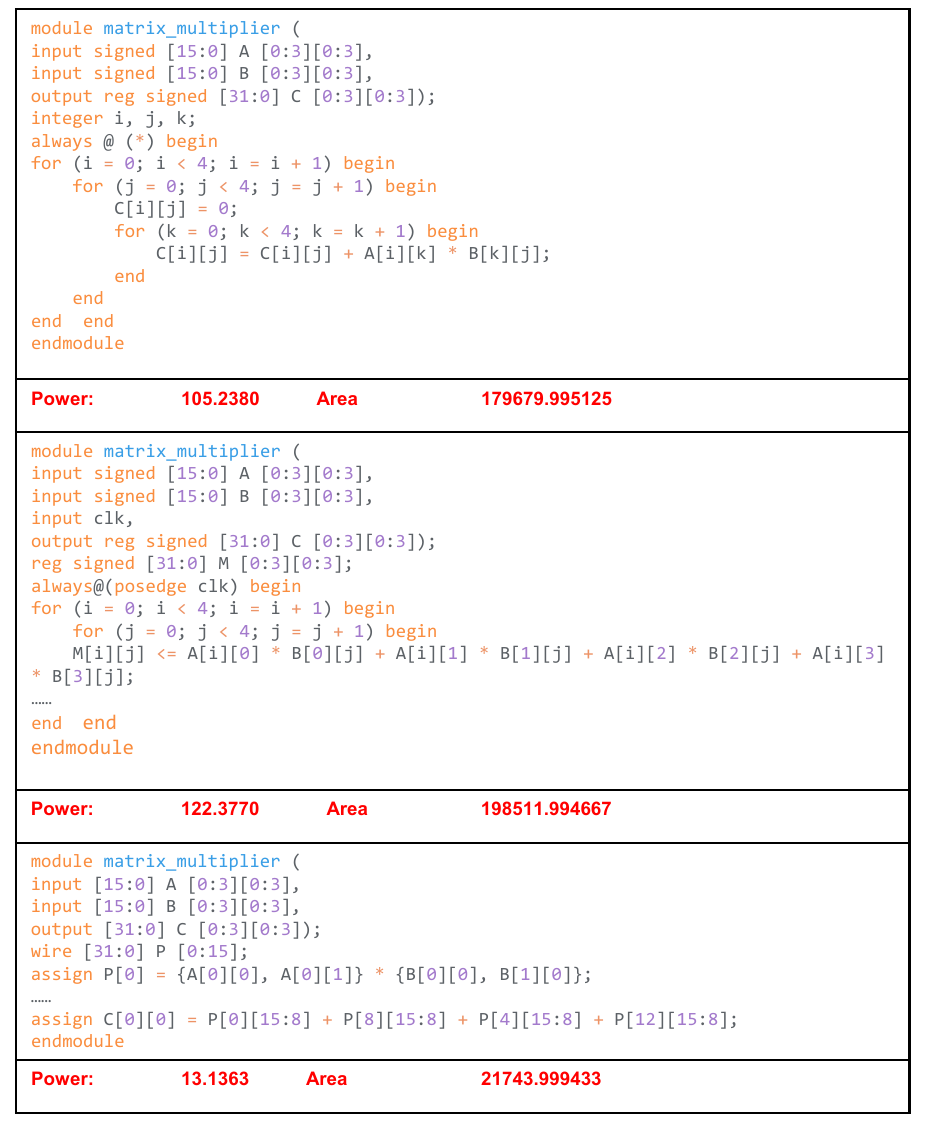} %调节单个子图大小
    \caption{Matrix Multiply Verilog Program list, where power is measured in milliwatts ($mW$), area is measured in micrometers squared (${\mu}m^2$)} %子图下标题
    \label{fig:matrix} %引用标签
\end{figure}
As discussed in Section \ref{sec:motivation}, power, performance, and area (PPA) are crucial metrics in chip logic design. However, GPT lacks comparable PPA configurations. In GPT model, the reward model only provides general ranking. Therefore, ChipGPT proposes a cost manager to optimize PPA following the GPT model. %insight

This stage uses design tools to test the PPA of each program, appending the
results to the program list (Equ. \ref{equ:plist}) until it completes an
enumerated search to select the target version. For example, in Fig.
\ref{fig:matrix}, GPT generates different versions, they have different PPAs.
Therefore, the cost manager tries to use EDA tools to output their PPAs and
search for the optimal one.%action
\subsection{Enumerative Search}

\begin{table}[]
    \label{tab:programrank}
    \caption{ChatGPT Program Ranking Model on button count description}
    \centering
    \begin{tabular}{ccccc}
        \rowcolor[HTML]{794DFF}
        \textcolor{white}{Program} & \begin{tabular}[c]{@{}c@{}}\textcolor{white}{Power}  \\ \textcolor{white}{($mW$)}\end{tabular} & \begin{tabular}[c]{@{}c@{}}\textcolor{white}{Area}  \\ \textcolor{white}{($\mu m^2$)}\end{tabular} & \begin{tabular}[c]{@{}c@{}}\textcolor{white}{Latency}\\\textcolor{white}{(Cycle)}\end{tabular} & \begin{tabular}[c]{@{}c@{}}\textcolor{white}{Rank}    \\ \textcolor{white}{Model}\end{tabular} \\
        \rowcolor[HTML]{ECE9FF}
        Raw   Program 0            & 4.2900e-02                                                                                     & 139.199999                                                                                         & 1                                                                                              & 5                                                                                              \\
        \rowcolor[HTML]{FFFFFF}
        Raw   Program 1            & 1.3593e-02                                                                                     & 265.200006                                                                                         & 1                                                                                              & 4                                                                                              \\
        \rowcolor[HTML]{ECE9FF}
        Raw   Program 2            & 1.0704e-02                                                                                     & 193.600004                                                                                         & 1                                                                                              & 3                                                                                              \\
        \rowcolor[HTML]{FFFFFF}
        Raw   Program 3            & 9.7253e-03                                                                                     & 187.200004                                                                                         & 1                                                                                              & 2                                                                                              \\
        \rowcolor[HTML]{ECE9FF}
        Raw   Program 4            & 1.0283e-02                                                                                     & 196.000003                                                                                         & 1                                                                                              & 1
    \end{tabular}
\end{table}
\begin{table}[htbp]
    \label{tab:searchresult}
    \caption{Search Result }
    \centering
    \begin{tabular}{llll}
        \rowcolor[HTML]{00BAC8}
        \begin{tabular}[c]{@{}l@{}} \textcolor{white}{Select by} \\ \textcolor{white}{Power}\end{tabular} & \begin{tabular}[c]{@{}l@{}}\textcolor{white}{Select by}\\  \textcolor{white}{Area}\end{tabular} & \begin{tabular}[c]{@{}l@{}}\textcolor{white}{Select by} \\ \textcolor{white}{Performance}\end{tabular} & \textcolor{white}{Direct Select} \\
        \rowcolor[HTML]{FFC000}
        P3                                                                                                & P0                                                                                              & P0                                                                                                     & P4                               \\
        \rowcolor[HTML]{E7F3F5}
        P4                                                                                                & P3                                                                                              & P1                                                                                                     & P1                               \\
        \rowcolor[HTML]{CBE7EB}
        P2                                                                                                & P2                                                                                              & P2                                                                                                     & P2                               \\
        \rowcolor[HTML]{E7F3F5}
        P1                                                                                                & P4                                                                                              & P3                                                                                                     & P3                               \\
        \rowcolor[HTML]{CBE7EB}
        P0                                                                                                & P1                                                                                              & P4                                                                                                     & P4
    \end{tabular}
\end{table}

The key question is whether the search stage is necessary. As shown in Table
\ref{tab:programrank}, the GPT model's ranking does not consistently match the
PPA rank. For example, searching the button count module programs produces the
results in Table \ref{tab:searchresult}. This shows that if we choose different
targets, the selected final program depends highly on the chosen target, as
highlighted in yellow. Therefore, adding a search algorithm to the output
manager is necessary. We use design space exploration to obtain better results
from the program list. The number of correct programs GPT generates is limited,
typically less than 10. As a result, we use enumerate search to select the
optimal result.

\begin{equation}\label{equ:plist}
    program\_list=\{\{p_1,PPA_1\},\dots,\{p_n,PPA_n\}\}
\end{equation}

where $p$ denotes a program. Enumerate search selects the best program in the
program list of performance, area and power, which relies on the programmers'
target.
%pareto optimization in PPA and reward model
%measure of PPA

\section{Evaluation}\label{sec:eval}
To explore the LLM ability in natural language hardware design and evaluate our
method's effectiveness, we list several key questions below(RQ) and try to
answer them with proofs.%the difficulty of evaluation

\paragraph{RQ1} Compare with other agile developing method(\emph{e.g.} Chisel, HLS), what
advantages do natural language based method have?

\paragraph{RQ2} Does our method(ChipGPT) takes effect compare with original ChatGPT in both PPA
results and programmability perspective?

\paragraph{RQ3} Does our framework sensitive to workloads? %是否需要测量code coverage有待考究

\paragraph{RQ4} Does the prompt principles (\emph{i.e.} ablation study of composition module,
interface model, and post-addition principle) really take effect to enhance the
soundness(\emph{i.e.} code quality) of our method?

\paragraph{RQ5} To automate the design flow, does the human feedback really takes a heavy
effect on our LLM-based framework?
\subsection{Experimental Setup}
\subsubsection{Experimental Environment}
We use the ChatGPT model from OpenAI's website, which is a GPT-3.5 model. The
raw programs of input are in SystemVerilog format. We use Design Compiler and
65nm technology to evaluate the power and area of output designs. The
performance cycle numbers of the designs produced through different approaches
are obtained by simulation with common handwritten testbeds. For line of code
measurement, the cloc tool is used.
\subsubsection{Programbility measurement}\label{sec:progmeas}

Traditional programmability evaluation in EDA uses lines of code (LOC) as the
measurement metric\cite{chisel}.
% For example, to evaluate Chisel programmability, researchers subtract the LOC of the generated Verilog from the Chisel program. 
This LOC subtraction method cannot directly measure natural language code
generation for two reasons: 1) The LOC of natural language is undefined. 2)
Natural language-based zero-code programming frameworks use neural networks,
which cannot ensure output programs strictly follow specifications. To address
these challenges, we define LOC subtraction as Equation \ref{equ:locmeasure} to
evaluate GPT code generation under the prompt manager: \vspace{-2mm} %limitation of traditional programmbility evaluation method
\begin{equation}\label{equ:locmeasure}
    quality=raw+correct-prompt
\end{equation}
Where $quality$ represents code generation algorithm efficiency, $raw$ is the number of lines in the generated program, $correct$ is the number of lines modified by humans to correct the raw program, and $prompt$ is the number of prompt queries.
\subsubsection{Benchmarks}
The benchmarks include several typical hardware structures and algorithm
accelerator implementations as shown in Table \ref{tab:workloaddetail}. Simple
CPU implements the instruction set in Tab. \ref{tab:cpuisa}. GPT exhibits
different performance on different workloads. Therefore, we classify them into
three parts based on complexity: 1) Composition (CM): The most complex
architecture, requiring module composition. 2) Complex single module (CSM):
More complex individual modules. 3) Simple single module (SSM): The simplest
modules, with identical program lists and PPA.
% For SSM workloads, the enumerated search is unnecessary because the program list contains only one program type with the same PPA. The output manager can select the module with the optimal PPA directly. For CM workloads, the module composition principle plays an important role. 
%workload的划分法，划分为三种benchmark，简单功能型、复杂功能型、组合型

\begin{table}
    \label{tab:workloaddetail}
    \centering
    \caption{workload detail}
    \begin{tabular}{l|l|l}
        Workload      & Type & Brief Illustration                                \\
        \hline
        matmul        & CSM  & Multiplying
        two matrices 4x4 with 16 bits                                            \\
        mux           & SSM  & A 4x1 Multiplexer                                 \\
        3-8decoder    & SSM  & Select one of the eight lines in a 3-to-8 decoder \\
        button        & CSM  & Counts
        the number of button presses                                             \\
        vecmat        & CSM  & Vector matrix multiply with 4-bits element        \\
        addmulti tree & CM   & Add multiply tree with 8-bit operand              \\
        accumulator   & CSM  & Sum an array of 8-bits elements                   \\
        simple CPU    & CM   & A simple CPU implementation
    \end{tabular}
\end{table}

% \usepackage[normalem]{ulem}
\begin{table*}[htbp]
    \centering
    \label{tab:cpuisa}
    \caption{Simple CPU ISA implementation}
    \begin{tabular}{|lllll|}
        \hline
        \multicolumn{1}{|l|}{\textbf{Instruction}} & \multicolumn{1}{l|}{\textbf{Description}}                                                & \multicolumn{1}{l|}{\textbf{Opcode}} & \multicolumn{1}{l|}{\textbf{Bit Function}}    & \textbf{Opcode   length(bit)} \\ \hline
        \multicolumn{5}{|c|}{Computing   Instructions}                                                                                                                                                                                                               \\ \hline
        \multicolumn{1}{|l|}{add}                  & \multicolumn{1}{l|}{Add reg{[}op2{]} to reg{[}op1{]}}                                    & \multicolumn{1}{l|}{00000}           & \multicolumn{1}{l|}{12b'x+5bop1+5b'op2+10b'x} & 32                            \\ \hline
        \multicolumn{1}{|l|}{mul}                  & \multicolumn{1}{l|}{Mul reg{[}op2{]} to reg{[}op1{]}}                                    & \multicolumn{1}{l|}{00001}           & \multicolumn{1}{l|}{12b'x+5bop1+5b'op2+10b'x} & 32                            \\ \hline
        \multicolumn{1}{|l|}{cmp}                  & \multicolumn{1}{l|}{Determine the equality of reg{[}op1{]} and reg{[}op2{]}.}            & \multicolumn{1}{l|}{00011}           & \multicolumn{1}{l|}{12b'x+5bop1+5b'op2+10b'x} & 32                            \\ \hline
        \multicolumn{5}{|c|}{Control  Instructions}                                                                                                                                                                                                                  \\ \hline
        \multicolumn{1}{|l|}{beq}                  & \multicolumn{1}{l|}{Determine if reg{[}op1{]} is greater than or equal to reg{[}op2{]}.} & \multicolumn{1}{l|}{00100}           & \multicolumn{1}{l|}{12b'x+5bop1+5b'op2+10b'x} & 32                            \\ \hline
        \multicolumn{1}{|l|}{ble}                  & \multicolumn{1}{l|}{Determine if reg{[}op1{]} is less than or equal to reg{[}op2{]}}     & \multicolumn{1}{l|}{00101}           & \multicolumn{1}{l|}{12b'x+5bop1+5b'op2+10b'x} & 32                            \\ \hline
        \multicolumn{1}{|l|}{blt}                  & \multicolumn{1}{l|}{Determine if reg{[}op1{]} is less than reg{[}op2{]}.}                & \multicolumn{1}{l|}{00110}           & \multicolumn{1}{l|}{12b'x+5bop1+5b'op2+10b'x} & 32                            \\ \hline
        \multicolumn{1}{|l|}{bge}                  & \multicolumn{1}{l|}{Determine if reg{[}op1{]} is greater than or equal to reg{[}op2{]}.} & \multicolumn{1}{l|}{00111}           & \multicolumn{1}{l|}{12b'x+5bop1+5b'op2+10b'x} & 32                            \\ \hline
        \multicolumn{1}{|l|}{br}                   & \multicolumn{1}{l|}{PC Jump  to reg{[}31:0{]}}                                           & \multicolumn{1}{l|}{00010}           & \multicolumn{1}{l|}{32b'M}                    & 32                            \\ \hline
        \multicolumn{5}{|c|}{Memory Instructions}                                                                                                                                                                                                                    \\ \hline
        \multicolumn{1}{|l|}{store}                & \multicolumn{1}{l|}{Store data\_in to reg{[}op1{]}.}                                     & \multicolumn{1}{l|}{01000}           & \multicolumn{1}{l|}{12b'x+5bop1+15b'x}        & 32                            \\ \hline
        \multicolumn{1}{|l|}{load}                 & \multicolumn{1}{l|}{Load the value from reg{[}op1{]}   into data\_out.}                  & \multicolumn{1}{l|}{01001}           & \multicolumn{1}{l|}{12b'x+5bop1+15b'x}        & 32                            \\ \hline
    \end{tabular}
\end{table*}
\subsubsection{Baseline}
To validate our workflow works with ChatGPT, we compare it to the baseline
ChatGPT model. The naive code ChatGPT generates uses only the module
description as the prompt. To analyze improvement over traditional agile
workflows, we compare with Chisel and high-level synthesis (Xilinx Vivado HLS).
\textbf{For a fair comparison, 1) we add unroll directive to pipeline the HLS
    design and used Synopsys Design Compiler to measure power and area, as with our
    approach. 2) The HLS and Chisel implementations follow the same specification
    and were developed by the same 2-year graduate student and evaluated with a
    coverage identical testbed suite.}
%The HLS compiler optimizer we used introduces area changes that are tangential to this experiment. While these optimizations impact the total area results, they do not affect the conclusions of this study.

% Because HLS and Chisel compilers involve trade-offs between area and latency, the measured PPA data may differ to some extent. However, the differences in power and area caused by these compiler optimizations do not affect the conclusions of the experiment. This is because the purpose of our experiment is to compare the PPA improvements of chipgpt and chatgpt, and then argue that using llm to generate verilog directly has lower latency. The research conclusions do not involve the changes in power and area caused by the trade-off between chisel and hls compilation.
%由于HLS和Chisel编译器的优化是一个在面积和延迟之间tradeoff的过程，会使得测量的PPA数据有一些差别。但是这种编译优化造成的Power和area差别并不影响实验结论。这是由于我们的实验目的是比较chipgpt和chatgpt的PPA提高，进而论证使用llm直接生成verilog的延迟较小，研究结论不涉及对chisel和hls编译tradeoff导致的power和area变化的。
%baseline introduction

\begin{table}[htbp]
    \label{tab:ppacomparasion}
    \centering
    \caption{PPA comparison to show the efficiency of our method under the area optimization target, where power is measured in milliwatts ($mW$), area is measured in micrometers squared (${\mu}m^2$) and latency measures the number of running cycles}
    \begin{tabular}{|c|l|l|l|r|}
        \hline
        \multicolumn{1}{|l|}{\textbf{Workload}}                                           & \textbf{Configuration} & \textbf{Power}                & \textbf{Area}                & \multicolumn{1}{l|}{\textbf{Latency}} \\ \hline
        \multirow{4}{*}{matrix mul}                                                       & ChatGPT(Baseline)      & 105.24                        & 179680.0                     & 1                                     \\ \cline{2-5}
                                                                                          & HLS                    & \multicolumn{1}{r|}{0.1946}   & \multicolumn{1}{r|}{2592.79} & 169                                   \\ \cline{2-5}
                                                                                          & Chisel                 & \multicolumn{1}{r|}{28.5361}  & \multicolumn{1}{r|}{55983.6} & 1                                     \\ \cline{2-5}
                                                                                          & Ours(ChipGPT)          & 13.14                         & 952.4                        & 1                                     \\ \hline
        \multirow{4}{*}{mux4x1}                                                           & ChatGPT(Baseline)      & \multicolumn{1}{r|}{2.62E-03} & \multicolumn{1}{r|}{11.2}    & 1                                     \\ \cline{2-5}
                                                                                          & HLS                    & \multicolumn{1}{r|}{2.62E-03} & \multicolumn{1}{r|}{11.2}    & 1                                     \\ \cline{2-5}
                                                                                          & Chisel                 & \multicolumn{1}{r|}{2.62E-03} & \multicolumn{1}{r|}{11.2}    & 1                                     \\ \cline{2-5}
                                                                                          & Ours(ChipGPT)          & \multicolumn{1}{r|}{2.62E-03} & \multicolumn{1}{r|}{11.2}    & 1                                     \\ \hline
        \multirow{4}{*}{3-8decoder}                                                       & ChatGPT(Baseline)      & \multicolumn{1}{r|}{2.60E-03} & \multicolumn{1}{r|}{22.8}    & 1                                     \\ \cline{2-5}
                                                                                          & HLS                    & \multicolumn{1}{r|}{7.03E-03} & \multicolumn{1}{r|}{156.4}   & 9                                     \\ \cline{2-5}
                                                                                          & Chisel                 & \multicolumn{1}{r|}{3.27E-03} & \multicolumn{1}{r|}{24.4}    & 1                                     \\ \cline{2-5}
                                                                                          & Ours(ChipGPT)          & \multicolumn{1}{r|}{2.60E-03} & \multicolumn{1}{r|}{22.8}    & 1                                     \\ \hline
        \multirow{4}{*}{button-count}                                                     & ChatGPT(Baseline)      & 0.01                          & 265.2                        & 1                                     \\ \cline{2-5}
                                                                                          & HLS                    & \multicolumn{1}{r|}{7.80E-03} & \multicolumn{1}{r|}{200.4}   & 9                                     \\ \cline{2-5}
                                                                                          & Chisel                 & \multicolumn{1}{r|}{8.34E-03} & \multicolumn{1}{r|}{146.8}   & 1                                     \\ \cline{2-5}
                                                                                          & Ours(ChipGPT)          & \multicolumn{1}{r|}{4.29E-02} & \multicolumn{1}{r|}{139.2}   & 1                                     \\ \hline
        \multirow{4}{*}{vector-matrix}                                                    & ChatGPT(Baseline)      & 1.30                          & 3451.2                       & 2                                     \\ \cline{2-5}
                                                                                          & HLS                    & 0.03                          & 428.8                        & 191                                   \\ \cline{2-5}
                                                                                          & Chisel                 & 1.27                          & 3400.0                       & 1                                     \\ \cline{2-5}
                                                                                          & Ours(ChipGPT)          & 1.15                          & 3144.0                       & 1                                     \\ \hline
        \multirow{4}{*}{\begin{tabular}[c]{@{}c@{}}adder-multi\\       tree\end{tabular}} & ChatGPT(Baseline)      & 28.50                         & 60070.4                      & 1                                     \\ \cline{2-5}
                                                                                          & HLS                    & 0.09                          & 1784.4                       & 73                                    \\ \cline{2-5}
                                                                                          & Chisel                 & 28.54                         & 55983.6                      & 1                                     \\ \cline{2-5}
                                                                                          & Ours(ChipGPT)          & 27.79                         & 50498.8                      & 1                                     \\ \hline
        \multirow{4}{*}{accumulator}                                                      & ChatGPT(Baseline)      & 0.02                          & 174.0                        & 1                                     \\ \cline{2-5}
                                                                                          & HLS                    & \multicolumn{1}{r|}{1.10E-02} & \multicolumn{1}{r|}{204.0}   & 17                                    \\ \cline{2-5}
                                                                                          & Chisel                 & \multicolumn{1}{r|}{2.57E-02} & \multicolumn{1}{r|}{136.4}   & 1                                     \\ \cline{2-5}
                                                                                          & Ours(ChipGPT)          & 0.03                          & 136.0                        & 1                                     \\ \hline
        \multirow{4}{*}{Simple CPU}                                                       & ChatGPT(Baseline)      & 2.57                          & 23138.4                      & 5                                     \\ \cline{2-5}
                                                                                          & HLS                    & 0.10                          & 2780.0                       & 38                                    \\ \cline{2-5}
                                                                                          & Chisel                 & 1.48                          & 25346.0                      & 3                                     \\ \cline{2-5}
                                                                                          & Ours(ChipGPT)          & 0.48                          & 3240.8                       & 5                                     \\ \hline
    \end{tabular}
\end{table}

\begin{figure}[htbp] %通栏

    \centering
    \includegraphics[width=1.0\linewidth]{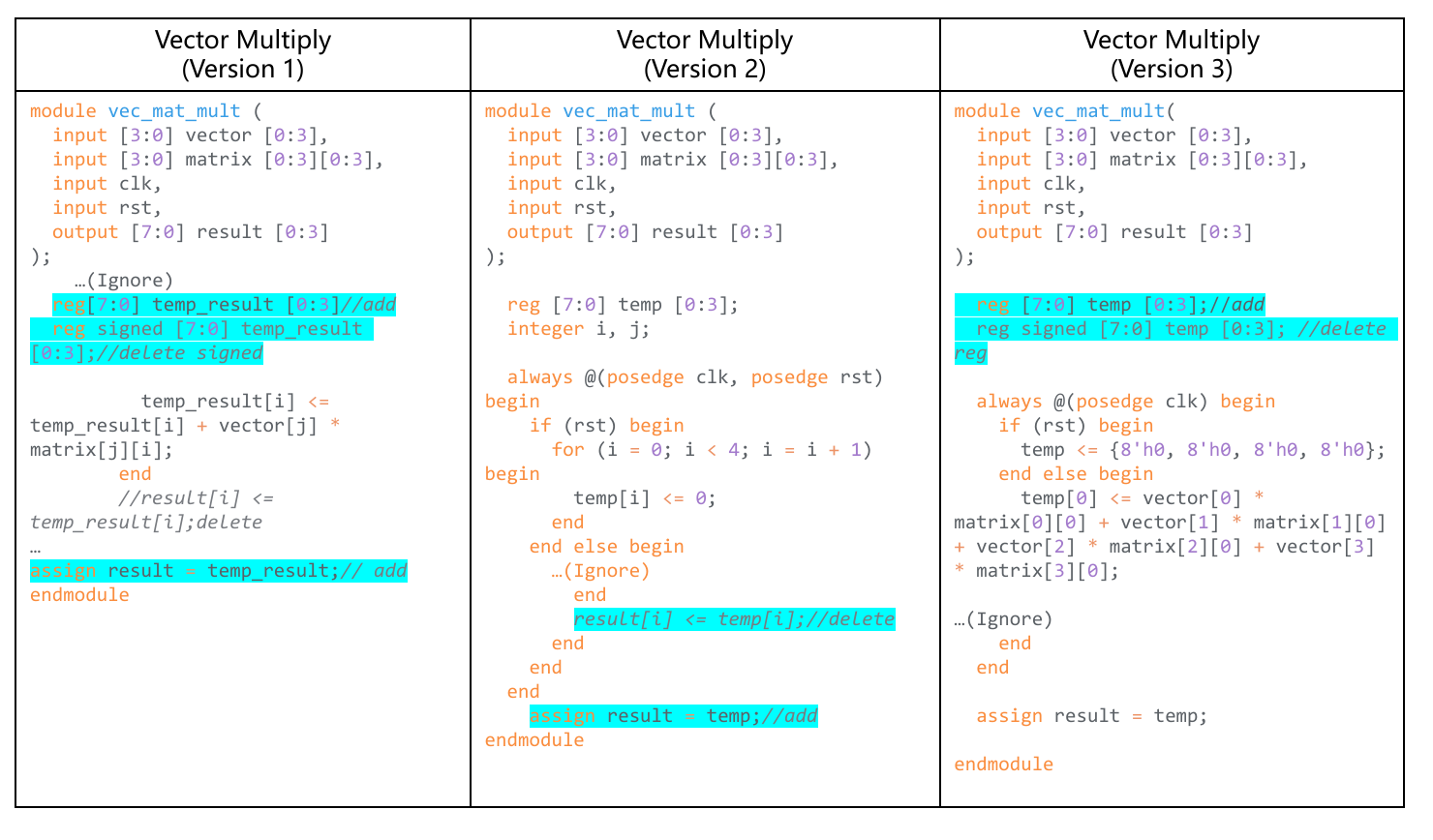} %调节单个子图大小
    \caption{Human Correct Implementation of Vector Multiply} %子图下标题
    \label{fig:correctcase} %引用标签
\end{figure}

\begin{figure*}

    \begin{minipage}[t]{0.33\linewidth} %调节两个子图左右间距
        \centering
        \centering
        \includegraphics[width=\linewidth]{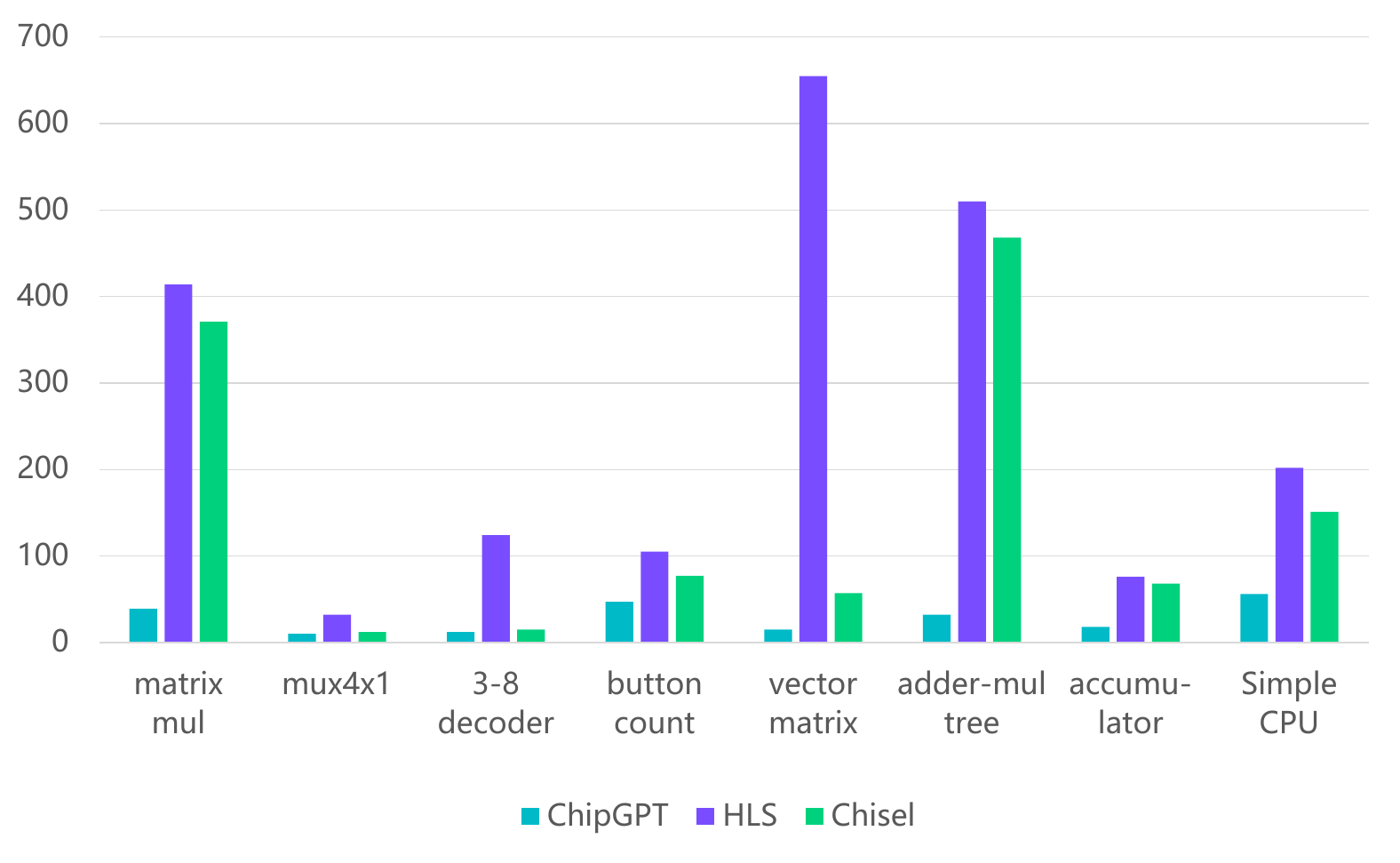}
        \caption{Generated code line number}
        \label{fig:rawcode}
    \end{minipage}
    \begin{minipage}[t]{0.33\linewidth} %调节两个子图左右间距
        \centering
        \includegraphics[width=\linewidth]{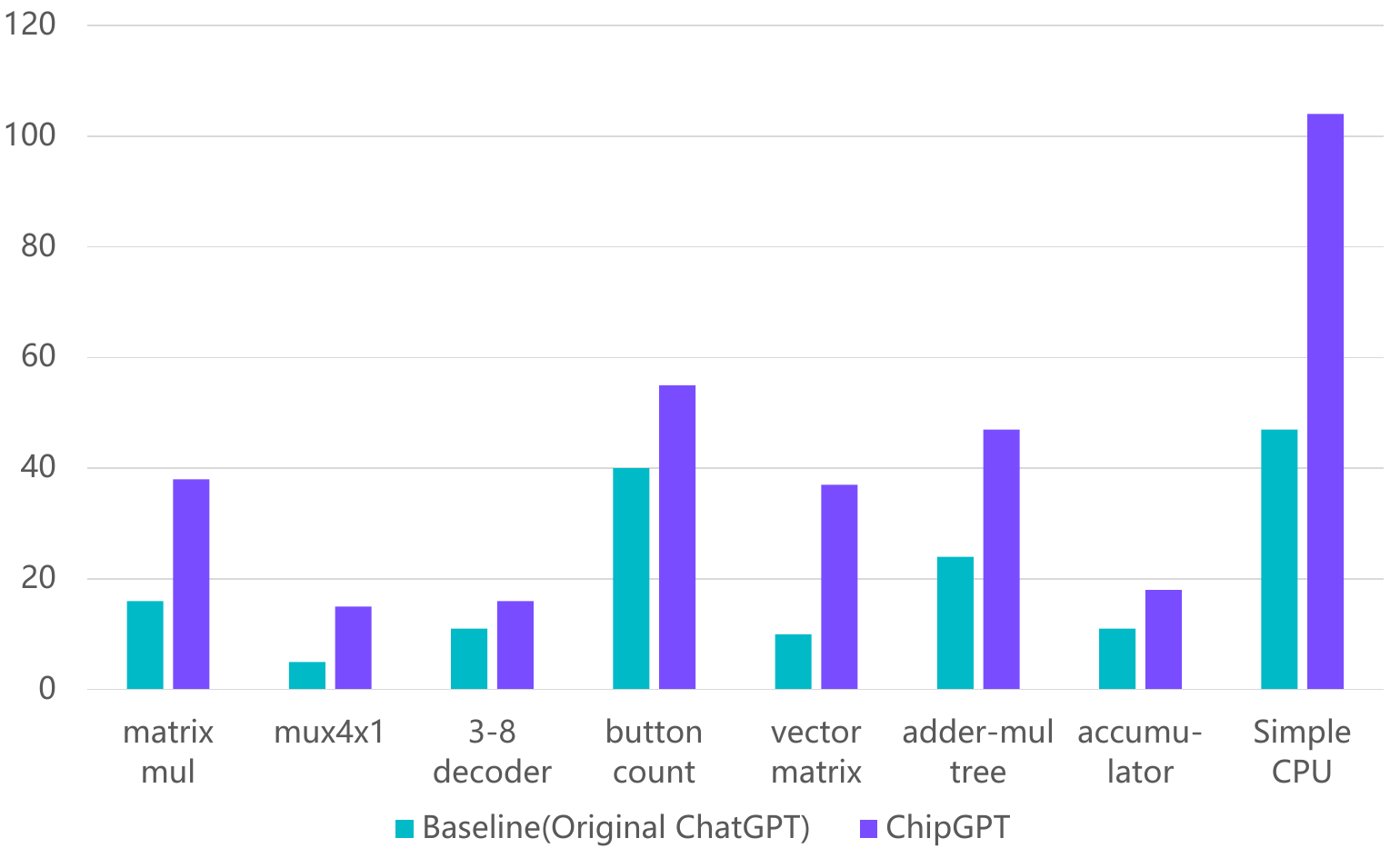}
        \caption{Code quality}
        \label{fig:codequality}
    \end{minipage}
    \begin{minipage}[t]{0.33\linewidth} %调节两个子图左右间距
        \centering
        \includegraphics[width=\linewidth]{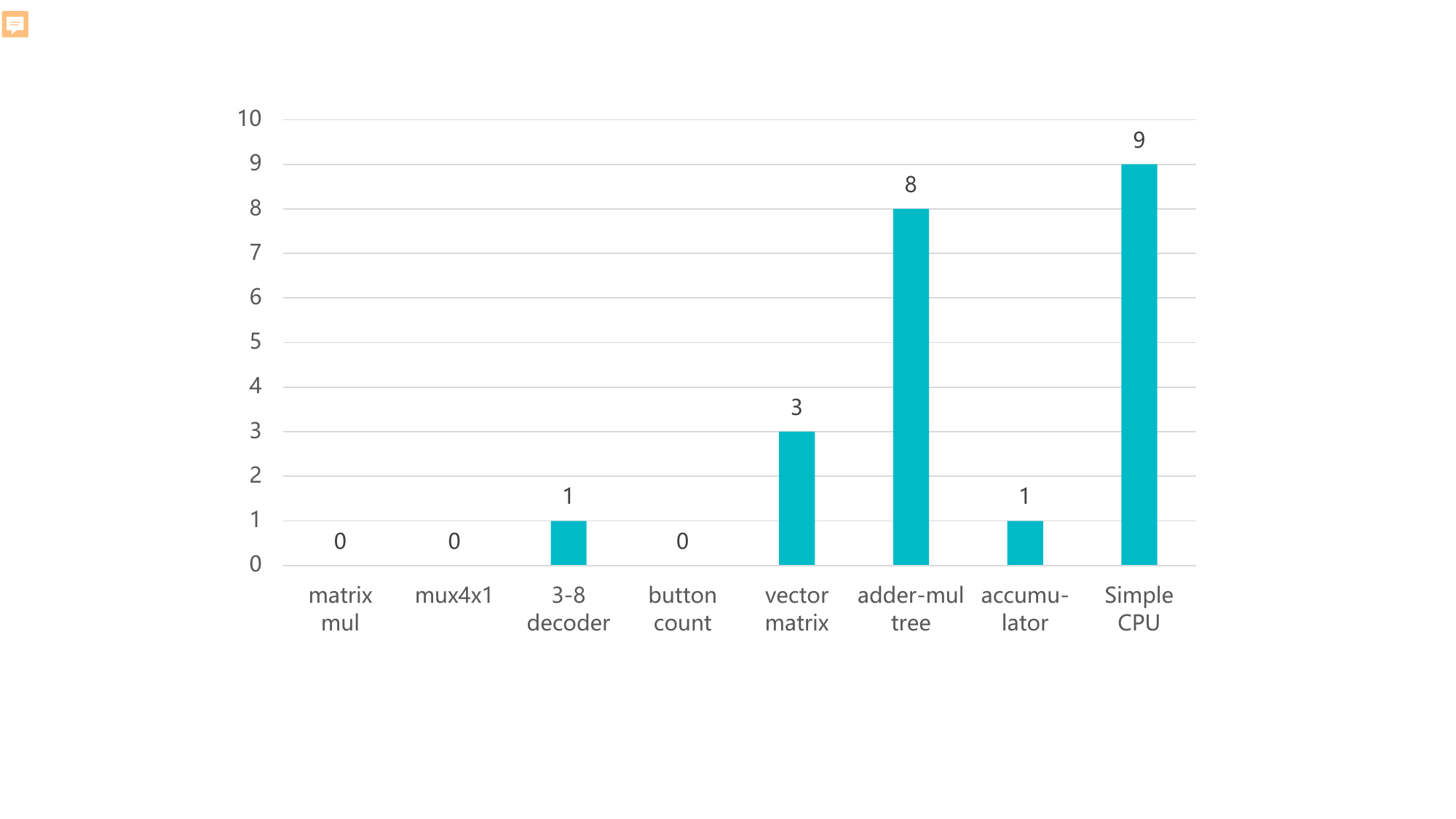}
        \caption{Human correction effort measured in line number of code}
        \label{fig:humanfeedback}
    \end{minipage}
\end{figure*}
\subsection{Experimental Results}

%basic table data 
To answer the above questions, we analyze our framework under different
workloads. Tab. \ref{tab:ppacomparasion} shows the power, performance, and area
(PPA) results compared to other agile methods when optimizing for the area. As
the HLS compiler prefers to use state machines rather than pipeline for
scheduling, sometimes resulting in areas smaller than the Chisel counterparts
and also longer latencies than that of Chisel. The static power consumption is
directly related to the total area. For example, in the matrix multiplication
workload, our HLS solution has lower area and power consumption compared to
Chisel, but exhibits much higher latency.
% The detailed experimental results are in \href{https://anonymous.4open.science/r/chipgptresult-C12E}{https://anonymous.4open.science/r/chipgptresult-C12E}.

We found several interesting findings during our generation process, which
helps us understand natural language digital design deeply. These findings and
insights provide answers to our five research questions.
\begin{framed}
    \vspace{-2mm}
    \textbf{Finding 1:} Natural language methods could achieve higher programmability than conventional agile hardware design techniques. (Answer RQ1)
    \vspace{-2mm}
\end{framed}
Evidence: Fig. \ref{fig:rawcode} compares the number of code lines generated for the same designs by our language model ChipGPT, high-level synthesis (HLS) tools, and the Chisel hardware design framework. On average, ChipGPT decreased the code volume by 9.25 times compared to HLS and 5.32 times compared to Chisel. This substantial reduction clearly demonstrates the enhanced programmability enabled by natural language techniques.

\begin{framed}
    \vspace{-2mm}
    \textbf{Finding 2:} Our four-stage framework optimizes PPA metrics while also improving code quality versus the original ChatGPT model. (Answer RQ2)
    \vspace{-2mm}
\end{framed}
Evidence: Fig. \ref{fig:codequality} illustrates the improved program generation efficiency enabled by our framework under the programmability metrics defined in Sec. \ref{sec:progmeas}. Maximum quality improvement corresponds to a 2.01 times reduction in the number of incorrect code lines required, demonstrating a more compact, higher-quality final design.
%由于自然语言范围搜索空间变大，因此搜到的PPA往往也更好
\begin{framed}
    \vspace{-2mm}
    \textbf{Finding 3:} Our framework improves program quality across workloads while also primarily optimizing PPA for complicated designs. On simple workloads, it shows limited PPA benefits due to the small pool of raw candidates and lack of opportunity for optimization. (Answer RQ3)
    \vspace{-2mm}
\end{framed}
Evidence: Fig. \ref{fig:codequality} demonstrates our framework brings program generation quality improvement to all the evaluated workloads. However, Tab. \ref{tab:ppacomparasion} shows that for simple (SSM) workload, PPA results exhibit no significant optimization relative to the baseline, as the limited number of raw program candidates provides little room for improvement. In contrast, for complex workloads like the CMS and CM tasks, our framework's targeted optimization at the search stage reduced average area by 47\% (0.53x) and overall average area by 35\% (0.65x) compared to the original ChatGPT model. By jointly optimizing PPA objectives and program coherence for complex designs with ample raw candidates, our framework achieved substantial optimization.
\begin{framed}
    \vspace{-2mm}
    \textbf{Finding 4:} Our composition, interface, and post-addition principles improve raw program coherence with natural language specifications, enabling high code quality at a large scale. (Answer RQ4)
    \vspace{-2mm}
\end{framed}
Evidence: Fig. \ref{fig:codequality} show that a 2.01$\times$ average increase in code quality is brought by the three principles in the prompt manager to our framework. They can enhance single-module generation quality and consistency for complex, large-scale designs.
\begin{framed}
    \vspace{-2mm}
    \textbf{Finding 5:} Our framework requires minimal human feedback for module generation and single large-scale modules due to the effect of prompt management principles. (Answer RQ5)
    \vspace{-2mm}
\end{framed}
Evidence: Fig. \ref{fig:humanfeedback} shows less than 10 lines of code needing correction for all workloads. Simple modules like the 4x1 multiplexer required no feedback, demonstrating autonomous generation. For complex modules and integrated accelerators up to 100 lines of code, only minor corrections were needed. For some cases, fewer than 10 lines of code need to be corrected. As an example, we analyze the vector matrix multiplication in Figure \ref{fig:correctcase}. The lines highlighted in blue indicate incorrect code that requires correction. The comments following these highlighted lines provide the appropriate fixes to yield the correct implementation. In each program, the number of highlighted lines that require corrections is fewer than three.

% \section{Future Work}
% Although this work tried applying LLMs to the automatic chip design field, there are several major challenges to address. First, LLM can only generate modules that are known as common knowledge in terms of hardware expertise. Using a private model to generate more specialized modules is difficult.  %private and public tradeoff v2
% Second, LLM lacks of data on novel hardware architecture such as PIM and quantum devices. %new hardware workflow
% Third, how to enhance LLM to be aware of the architectural or software-hardware co-design is an important challenge. %higher level module
% Fourth, it has several difficulties when designing ultra-large modules. %ultra large module
\section{Conclusion}%//has been refined by claude
This paper explores natural language hardware design and proposes ChipGPT, a framework that uses natural language specifications for automatic chip logic design. It  integrates language models into EDA tools without retraining. This could enable natural language chip design flows, reducing code volume by 5.32-9.25 compared to traditional agile methods.
By harnessing language, ChipGPT significantly accelerates chip development. ChipGPT is an interface for GPT to address natural language hardware design and PPA optimization, which has an area reduction of 47\% compared with the original ChatGPT in area target optimization mode. It also improves the correctness of LLM from probable right to rule right.
%more insightful conclusion

% \section*{Acknowledgment}

\bibliography{foo}
\bibliographystyle{ieeetr}

\vspace{12pt}

%fig:语言表达能力结构图（三维坐标）
%fig:不同种类程序合成技术对比表格
%fig:motivation的prompt输出图
%fig:对应每个principle实践的原语列表（实验列出）
\end{document}